\documentclass[journal]{IEEEtran}
\usepackage{times}
\usepackage{epsfig}
\usepackage{graphicx}
\usepackage{amsmath}
\usepackage{amssymb}
\usepackage{multirow}
\usepackage{array}
\usepackage{arydshln}
\usepackage{booktabs}
\usepackage[table]{xcolor}
\usepackage{amsmath}
\usepackage{subfigure}
\usepackage{enumerate}
\usepackage{adjustbox}
\usepackage{amsfonts}
\usepackage{fdsymbol}
\usepackage{color,xcolor}
\usepackage{colortbl}
\usepackage{comment}
\definecolor{mygray}{gray}{.9}

\newtheorem{lemma}{Lemma}[section]
\newtheorem{proof}{Proof}[section]

\newcommand{\myref}[1]{Eq.\eqref{#1}}
\newcommand{\etc}[0]{\textit{etc.}}
\newcommand{\ie}[0]{\textit{i.e.,}}
\newcommand{\eg}[0]{\textit{e.g.,}}
\newcommand{\etal}[0]{\textit{et al.}}

\ifCLASSINFOpdf
\else
\fi
\begin{document}
%
\title{Learning Domain Invariant Representations for Generalizable Person Re-Identification}%
%
%

\author{Yi-Fan Zhang, Zhang Zhang, Da Li, Zhen Jia, \\Liang Wang,~\IEEEmembership{Fellow,~IEEE}
        and~Tieniu Tan,~\IEEEmembership{Fellow,~IEEE}
\thanks{Yi-Fan Zhang, Zhang Zhang, Da Li, and Zhen Jia are with the National Laboratory of Pattern Recognition (NLPR), Center for Research on Intelligent Perception and Computing (CRIPAC), Institute of Automation, Chinese Academy of Sciences (CASIA), Beijing 100190, China, and also with the School of Artificial Intelligence, University of Chinese Academy of Sciences (UCAS), Beijing 100049, China (e-mail: yifanzhang.cs@gmail.com; zzhang@nlpr.ia.ac.cn, da.li@cripac.ia.ac.cn, zhen.jia@nlpr.ia.ac.cn).}
\thanks{Liang Wang and Tieniu Tan are with the National Laboratory of PatternRecognition, Center for Research on Intelligent Perception and Computing,Institute of Automation, Chinese Academy of Sciences (CASIA), Beijing100190, China, also with the University of Chinese Academy of Sciences,Beijing  100044,  China,  and  also  with  the  Center  for  Excellence  inBrain  Science  and  Intelligence  Technology,  Institute  of  Automation,Chinese Academy of Sciences (CASIA), Beijing 100190, China (e-mail:wangliang@nlpr.ia.ac.cn; tnt@nlpr.ia.ac.cn)}
\thanks{Corresponding author: Zhang Zhang.}
}

%
%

\markboth{Journal of \LaTeX\ Class Files,~Vol.~14, No.~8, August~2015}%
{Shell \MakeLowercase{\textit{et al.}}: Bare Demo of IEEEtran.cls for IEEE Journals}
%



\maketitle

\begin{abstract}
Generalizable person Re-Identification (ReID) aims at learning ready-to-use cross-domain representations for direct cross-data evaluation, which has attracted growing attention in the recent computer vision (CV) community. In this work, we construct a structural causal model (SCM) among identity labels, identity-specific factors (clothing/shoes color \etc), and domain-specific factors (background, viewpoints \etc). According to the causal analysis, we propose a novel Domain Invariant Representation Learning for generalizable person Re-Identification (DIR-ReID) framework. Specifically, we propose to disentangle the identity-specific and domain-specific factors into two independent feature spaces, based on which an effective backdoor adjustment approximate implementation is proposed for serving as a causal intervention towards the SCM. Extensive experiments have been conducted, showing that DIR-ReID outperforms state-of-the-art (SOTA) methods on large-scale domain generalization (DG) ReID benchmarks.
\end{abstract}


%
\IEEEpeerreviewmaketitle

\section{Introduction}
%
%
%
%
\begin{figure}[h]
  \begin{center}
   \includegraphics[width=1.0\linewidth,scale=0.9]{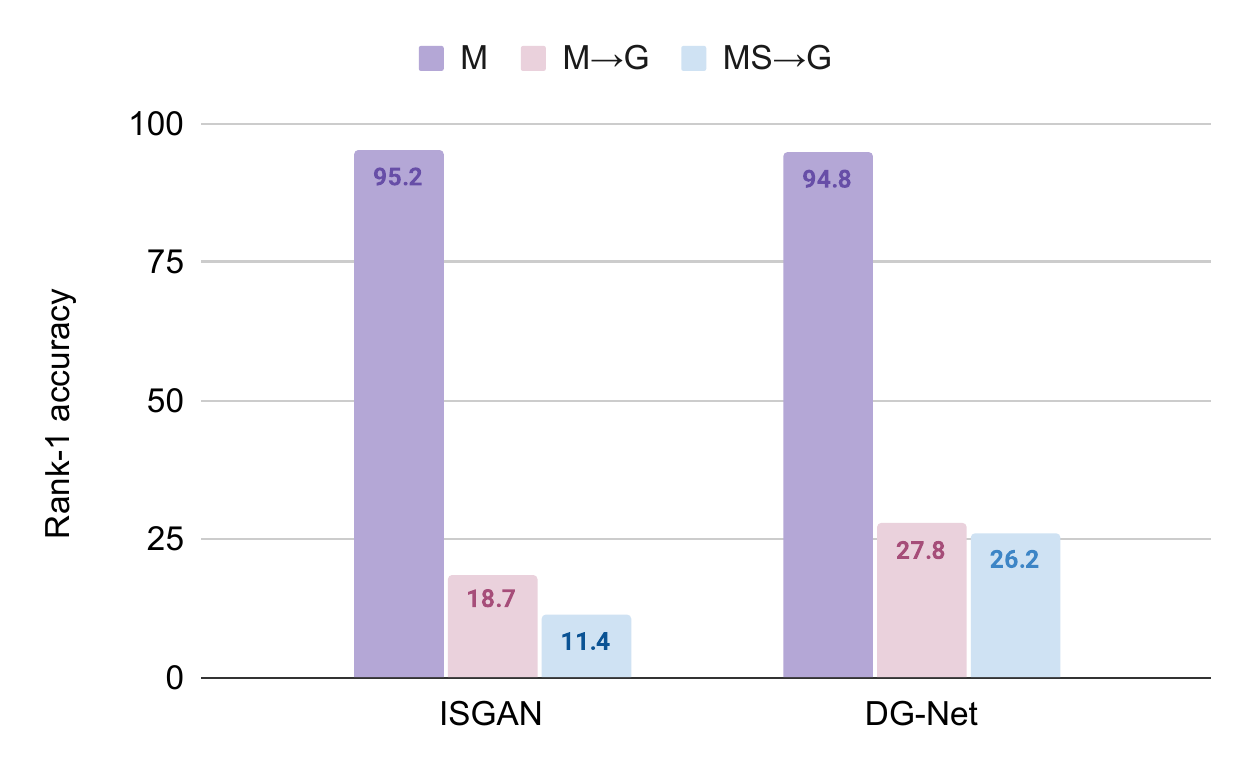}
  \end{center}
   \caption{Performance of two representative models. M: Train and test on Market1501. M$\rightarrow$G: Trained on Market1501 and tested on GRID . MS$\rightarrow$G: Trained on multi-source datasets and tested on GRID.}
  \label{fig:compare_models}
\end{figure}
Person Re-IDentification (ReID)~\cite{ye2021deep} aims at matching person images of the same identity across multiple camera views. In previous work, most ReID models are trained and tested on the same dataset, termed fully-supervised methods~\cite{Zhang_2020_CVPR,Zheng_2019_CVPR}, or adapted by unlabeled data in target domains different form the training datasets, termed unsupervised domain adaptation (UDA) methods~\cite{luo2020generalizing,Qi_2019_ICCV}. Although recent fully-supervised methods have achieved remarkable performance, they tend to fail catastrophically when tested in out-of-distribution (OOD) settings. \figurename~\ref{fig:compare_models} illustrates the fragility of two representative fully-supervised models, \ie, the DG-Net~\cite{Zheng_2019_CVPR} and ISGAN~\cite{ISGAN}, which both get very high rank-1 accuracies when model training and test are performed on the same Market1501 dataset~\cite{Zheng_2015_ICCV}. However, the rank-1 accuracies drop to $18.7\%$ and $27.8\%$ respectively when directly tested on the GRID dataset~\cite{grid}, suggesting the weak extrapolation capability and poor robustness of fully supervised methods. We further train these two models over multiple source domains (the details of sources are in Section~\ref{sec:dataset}). However, \textit{the even worse performance are obtained}, which indicates the challenge of ReID in the OOD settings. To tackle these problems, a number of UDA methods have been proposed to mitigate the domain gap without the need for extensive manual annotations in new target domains. However, they still need to collect large amounts of unlabeled data for UDA retraining. These problems severely hinder real-world applications of current person ReID techniques.

To tackle the above challenges, we focus on a more realistic and practical setting: generalizable person ReID, where the model is trained on multiple large-scale public datasets and on unseen domains directly without any model adaptations. The generalizable person ReID is originally formulated as a problem of domain generalization (DG)~\cite{Song_2019_CVPR}, which is more practicable than the traditional ReID paradigm since the ready-to-use models can work on any new settings without the requirement of data collection, annotation, and model updating. 

Assuming that a person image can be factorized into two latent factors, \ie, the identity-specific factors $S$ (\eg~appearances, body shapes) and the domain-specific factors $V$ (\eg~imaging conditions such as backgrounds, viewing angles, illuminations \etc ), we first present a structural causal model (SCM) for generalizable person ReID, which provides insights for the poor generalization of traditional ReID models when applied to unseen domains. Here, we highlight the potential reason for poor generalization ability: the domain-specific factors $V$ confound the identity-specific factors $S$ as well as the identity label $Y$, so that the spurious correlations between $V$ and $S$ hinder the model from making a robust prediction on identity label $Y$ based on $S$. Thus, a novel domain-invariant representation learning paradigm is proposed for generalizable person ReID, termed DIR-ReID, which disentangles the two latent factors $V$ and $S$ to remove spurious features.

Specially, a Multi-Domain Disentangled Adversarial Neural Network (MDDAN) is first proposed to jointly learn two encoders for embedding identity-specific and domain-specific factors from multiple public ReID datasets, where the adversarial learning principle is adopted to exclude the domain-related information from the embedded identity specific representations. Then a differentiable backdoor adjustment block (BA) is proposed to approximate the interventional distribution~\cite{neuberg2003causality}, which can pursue the true causality between identity-specific factors and identity labels. These two components (MDDAN and BA) are integrated as an end-to-end learning framework for generalizable person ReID, namely DIR-ReID. 

To sum up, the contributions of our work can be summarized as follows: 
\begin{itemize}
  \item For the first time, a causal perspective on the analysis of generalizable person ReID is introduced, by which the domain-specific factor is essentially a confounder that causes spurious correlations between person features and identity labels in new target domains.
  \item Thanks to the above analysis, a novel domain-invariant representation learning framework is proposed for generalizable person ReID, namely DIR-ReID, where an MDDAN block is adopted to disentangle identity-specific and domain-specific factors from multiple data sources. Then a BA block is adopted to approximate causal interventions. Mathematical analysis proves the characteristics of our method; 
  \item Comprehensive experiments are conducted to demonstrate the effectiveness of the proposed DIR-ReID model. Our method achieves superior performance in comparison with state-of-the-art (SOTA) methods on large-scale generalizable person ReID benchmarks.
\end{itemize}

\section{Related Work}
 {
\textbf{Single-domain Person ReID}
Existing works of single-domain person ReID (\ie~supervised person ReID) usually depend on the assumption that training and testing data are independent and identically distributed. They usually design to learn discriminative features \cite{matsukawa2016hierarchical} or develop efficient metrics \cite{koestinger2012large}. With the rapid development of deep Convolutional Neural Networks (CNNs), single-domain person ReID has achieved great progress. Some of the CNN-based methods introduce human parts \cite{li2017learning,pcb}, poses \cite{zhang2019densely}, and masks \cite{song2018mask} to improve the robustness of extracted features.~\cite{wu2021person} propose a multi-level Context-aware Part Attention(CPA) model to learn discriminative and robust local part features.~\cite{ye2020visible} propose a Homogeneous Augmented Tri-Modal (HAT) learning method for visible modality and night-time infrared modality.~\cite{ye2021collaborative} introduces an online co-refining (CORE) framework with dynamic mutual learning.~\cite{ye2021dynamic} designs an intra-modality weighted-part attention (IWPA) to construct part-aggregated representation. And some other methods use deep metric learning to learn appropriate similarity measures \cite{chen2017beyond}. Due to the space limitation, many important works cannot be covered. A well-summarized survey on person reID can be found at~\cite{ye2021deep}. Despite the encouraging performance under the single-domain setup, current fully-supervised ReID models degrade significantly when deployed to an unseen domain.}

\textbf{Cross-domain Person ReID}
Unsupervised Domain Adaptation (UDA) technologies have great progress~\cite{peng2020d2v} and been widely adopted for cross-domain person ReID. The UDA-based ReID methods usually attempt to transfer the knowledge learned from the labeled source domains to target domains one depending on target-domain images \cite{luo2020generalizing,huang2020real}, features \cite{wang2018transferable} or metrics \cite{Peng_2016_CVPR}. Another group of UDA-based methods \cite{ge2020mutual,zhai2020multiple} propose to explore hard or soft pseudo labels in the unlabeled target domain using its data distribution geometry. Though UDA-based methods improve the performance of cross-domain ReID to a certain extent, most of them require a large amount of unlabeled target data for model retraining. 

\textbf{Generalizable Person ReID} Recently, generalizable person ReID methods \cite{Song_2019_CVPR} are proposed to learn a model that can generalize to unseen domains without the requirement of model adaptation and data collection in target domains. Existing methods mainly follow a meta-learning pipeline or utilize domain-specific heuristics. Jia \etal \cite{jia2019frustratingly} learn the domain-invariant features by integrating Instance Normalization (IN) into the network to filter out style factors. Jin \etal \cite{jin2020style} extend the work \cite{jia2019frustratingly} by restituting the identity-relevant information to network to ensure the model discrimination. Lin \etal \cite{lin2020multi} propose a feature generalization mechanism by integrating the adversarial auto-encoder and Maximum Mean Discrepancy (MMD) alignment. Song \etal \cite{Song_2019_CVPR} propose a Domain-Invariant Mapping Network (DIMN) following the meta-learning pipeline. There also have some studies for learning domain-invariant features, \eg~DANN~\cite{ganin2016domain}, DDAN~\cite{chen2020dual} and CaNE~\cite{yuan2020calibrated}. The difference between DIR-ReID and these methods is detailed in Section~\ref{sec:model_analysis}.

\textbf{Domain Generalization}
In the machine learning community, domain/OOD generalization \cite{gulrajani2020search,zhang2022towards,zhang2022domain,zhang2022generalizable} aims to learn representations $\Phi(X)$ that is invariant across environments $\mathcal{E}$ so that model can well extrapolate in unseen environments. The problem can be formulated as $\min_{\Phi} \max _{e \in \mathcal{E}} \mathbb{E}[l(y, \Phi(x))\mid E=e]$. Representative approaches such as IRM \cite{arjovsky2020invariant} have been proposed to address this challenge. However, the IRM would fail catastrophically unless the test data are sufficiently similar to the training distribution \cite{rosenfeld2020risks}. To alleviate these challenges, we adopt a causal representation~\cite{scholkopf2021towards} framework termed DIR-ReID, to explicitly remove the confounding effects of spurious features via backdoor adjustment.

\textbf{Causality for CV} Causal Representation Learning~\cite{scholkopf2021towards} combines machine learning and causal inference and has attracted increasing attention within a learning paradigm to improve generalization and trustworthiness. 
Simultaneously, there is a growing number of CV tasks that benefit from causality~\cite{atzmon2020causal,magliacane2018domain,zhang2022exploring}. Most of them focus on measuring causal effects: disentangling the desired model effects~\cite{besserve2019counterfactuals}, and modularizing reusable features that generalize well~\cite{parascandolo2018learning}. Recently, causal intervention is also introduced into some CV researches~\cite{wang2020visual,yang2020deconfounded,tang2020longtailed}. Specifically, CONTA~\cite{dong_2020_conta} removes the confounding bias in image-level classification by backdoor adjustment and thus provides better pseudo-masks as ground truth for optimizing the subsequent segmentation model. IFSL~\cite{yue2020interventional} believes that pre-training is a confounding factor that hurts the performance of few-shot learning (FSL). Thus, they propose an SCM in the FSL process and then develop three practical implementations based on the backdoor adjustment. We also adopt the SCM~\cite{pearl2000models} to model causal effects in generalizable person ReID, where the causal analysis clearly provides the explanations why traditional methods work poorly in unseen domains and then guides the design of the proposed DIR-ReID framework.

\section{Learning Disentangled and Invariant Representations}
In this section, we first introduce the proposed SCM to analyze spurious correlations between domain-specific factors $V$ and identity labels $Y$. Then, a DIR-ReID framework is proposed to learn domain-invariant features for generalizable person ReID, where a BA block approximates the interventional distribution to capture the true causality between identity-specific factors $S$ and identity labels $Y$. Finally, a theoretical analysis is taken for a better understanding of our method.

\subsection{SCM for Generalizable Person ReID}\label{sec:scm}
  year={2022}
Inspired by current research on harnessing causality in machine learning~\cite{atzmon2020causal}, we propose an SCM to analyze disentanglement and generalization in persegbibon ReID models. 
\begin{figure}[h]
  \begin{center}
  \subfigure[]{
  \begin{minipage}[t]{0.45\linewidth}
   \includegraphics[scale=0.5]{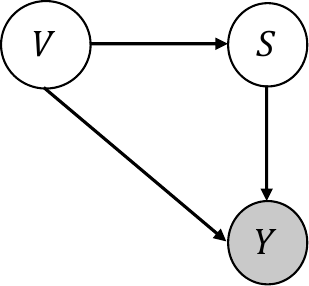}
   \label{fig:scm_reid}
   \end{minipage}}
   \subfigure[]{
   \begin{minipage}[t]{0.45\linewidth}
   \includegraphics[scale=0.5]{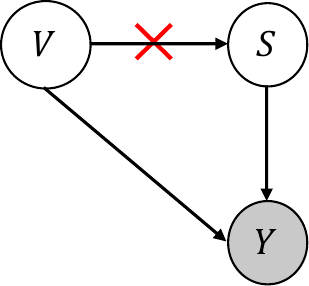}
   \label{fig:scm_causal}
   \end{minipage}}
  \end{center}
   \caption{Graphical representation of person ReID methods. $S,V$: identity-specific and domain-specific factors, $Y$: identity labels. Gray circles denote observable variables. (a) Traditional ReID model where $P(Y|S)\neq P(Y|do(S))$. (b) Interventional ReID model where $P(Y|S)\approx P(Y|do(S))$.}
  \label{fig:scm}
\end{figure}

Following the causal models in~\cite{cai2019learning,sun2020latent}, we use the SCM (in \figurename~\ref{fig:scm_reid}) to describe the causal relationships between person images and person identities, where $Y$  denotes the observable variables of identity labels, $S$ and $V$ are the latent variables indicating identity-specific and domain-specific factors, respectively. As shown in the model, there are three kinds of causal relationships as follows.

\noindent${S\rightarrow Y}.$ Identity-specific factors $S$ directly cause $Y$, which means that person identities are mainly determined by their identity-specific information, such as clothing styles, body shapes, \etc.

\begin{figure}[h]
  \begin{center}
   \includegraphics[width=0.8\linewidth,scale=0.3]{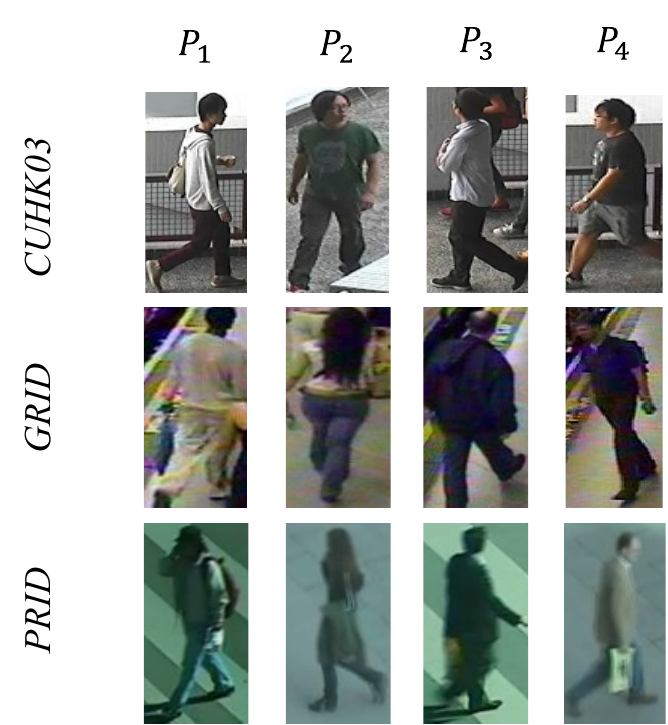}
  \end{center}
   \caption{Spurious correlations between domain-specific factors and identity-specific factors.}
  \label{fig:causal_imgs}
\end{figure}

\noindent${V\rightarrow S,V\rightarrow Y}.$ In real scenarios, there are also some confounders $v_i\in V$ (\eg, backgrounds, illuminations and viewpoints) that affect both identity factors $S$ and person identities $Y$. The $V\rightarrow S$ edge indicates the spurious correlations between $S$ and $V$ in the real world. $V\rightarrow Y$ denotes the influences of contextual environments in $V$ on $Y$. For example, most pedestrians in CUHK03 dataset are captured by high-definition cameras on the campus. Although for some early ReID datasets, \eg, GRID~\cite{grid}  and PRID~\cite{prid}, the low resolution, varying illumination conditions, and various parameters of imaging devices (domain-specific factors) make the appearance of persons vary greatly, even for the same color clothes. For example, the pedestrians in column 1 of \figurename~\ref{fig:causal_imgs} are all wearing white, however, their appearance, such as clothing colors (identity -specific factors) will be influenced by some confounders in the domain-specific factors\footnote{The white cloth of $P_1$ in GRID is more yellowish and that of $P_1$ in PRID is more greenish}. Thus, there are spurious correlations between $S$ and $V$ and $V$ also confounds the prediction of identity labels $Y$.

An ideal ReID model should capture the true causality between $S$ and $Y$, which can be generalized to other unseen domains. However, from the SCM in~\figurename~\ref{fig:scm_reid}, the conventional correlation $P(Y|S)$ fails to do so, because the likelihood of $Y$ given $S$ is not only influenced by ``S causes Y" via $S\rightarrow Y$, but also spurious correlations via $V\rightarrow Y$. Therefore, to pursue the true causality between $S$ and $Y$, we need to adopt the causal intervention~\cite{neuberg2003causality} $P(Y|do(S))$ instead of the likelihood $P(Y|S)$ for the ReID model.

\subsection{Preliminaries of Causal Intervention}

 The causal intervention seeks the true causal effect of one variable on another, and it is appealing for the objective of DG ReID: given one pedestrian image $x_i$, we hope the model prediction (pedestrian entity) is faithful only to the semantic feature $S$, while removing the effects of spurious associations from domain-specific factors $V$. We use~\figurename~\ref{fig:scm} for example, where domain-specific factors $V$ (\eg~imaging conditions such as backgrounds, viewing angles, illuminations \etc) affect both $S$ and $Y$, leading to spurious correlations if only learning from $P(Y|S)$. To see this, by using the Bayes rule: 

\begin{equation}
    P(Y|S)=\sum_{v}P(Y|S,v)P(v|S),
    \label{equ:bayes}
\end{equation}
where $v$ is an instance in $V$, which introduces the observational bias. Referring to the analyses in ~\cite{wang2020visual,yang2020deconfounded}, we assume that the training dataset contains much more pedestrian identities from the CUHK03 dataset, where $S$ denotes the semantic information of a pedestrian and $v_{c3}\in V$ the domain-specific factors of the CUHK03 dataset. After training, $P(v_{c3}|S) \approx 1$, hence $P(Y|S)$ degrades to $P(Y|S,v_{c3})$, which the supervised ReID methods actually do. Thus, conventional ReID methods tend to build strong connections between domain-specific factors and pedestrian identities in one domain, where the ReID model is contaminated by the backdoor path $S\leftarrow V\rightarrow Y$.

Do-operation~\cite{pearl2000models} removes certain relationships in the causal graph and replaces a factor with a constant. In our setting, the dependency between $V$ and $S$ should be cut off and the intervention posterior $P(Y|do(S))$ by applying do-operation will be:

\begin{equation}
    P(Y|do(S))=\sum_{v}P(Y|S,v)P(v).
\end{equation}
Compared to \myref{equ:bayes}, the key difference is that the adjustment weight $P(v|S)$ is changed to $P(v)$ because $V$ is no longer dependent on $S$. After the intervention, the cur-off (\figurename~\ref{fig:scm_causal}). This encourages DG-ReID models to maximize $P(Y|S,v)$ for every style factor $v$, only subject to a prior $P(v)$.

\subsection{Causal Intervention via Backdoor Adjustment}\label{sec:intervention}
The above formulation only gives a causal quantity $P(Y|do(S))$ without further \textit{identification} or grounding methods for computing it from purely statistical quantities. Therefore, we propose to use the backdoor adjustment~\cite{pearl2016causal} to identify and compute $P(Y|do(S))$ without the need for an ideal dataset\footnote{An ideal dataset has images of every pedestrian in all backgrounds and illumination conditions, which is unbiased and has few spurious correlations. Namely, there is no edge $V\rightarrow S$ and then $P(Y|S)=P(Y|do(S))$.}. The back-door adjustment assumes that we can stratify the analysis by a number of confounding factors, \ie~$V=\{v_i\}_{i=1}^{|V|}$, where each $v_i$ is the domain-specific factor corresponding to a certain camera view, illumination condition, \etc. Formally, the backdoor adjustment for the graph in~\figurename~\ref{fig:scm_causal} is (the detailed proof is shown in Appendix~\ref{proof:intervention}):

\begin{equation}
  \begin{aligned}
    P(Y|do(S=s_k))=&\sum_{i=1}^{|V|}P(Y|S=s_k,V=v_i)P(V=v_i)
  \end{aligned}
  \label{equ:intervention}
\end{equation}

To calculate the above intervention distribution, there are still two challenges: (i) it is hard to instantiate $v$ and $s$, namely learning two embedding functions, $v_i = f_v(x_i)$ and $s_i = f_s(x_i)$ where $x_i$ is the $i-$th person image. (ii) it is almost impossible to enumerate all domain-specific factors $V$. Next, we will offer a practical implementation of Eq.\eqref{equ:intervention}: DIR-ReID.

\begin{figure*}[t]
  \begin{center}
   \includegraphics[width=0.8\linewidth,scale=0.5]{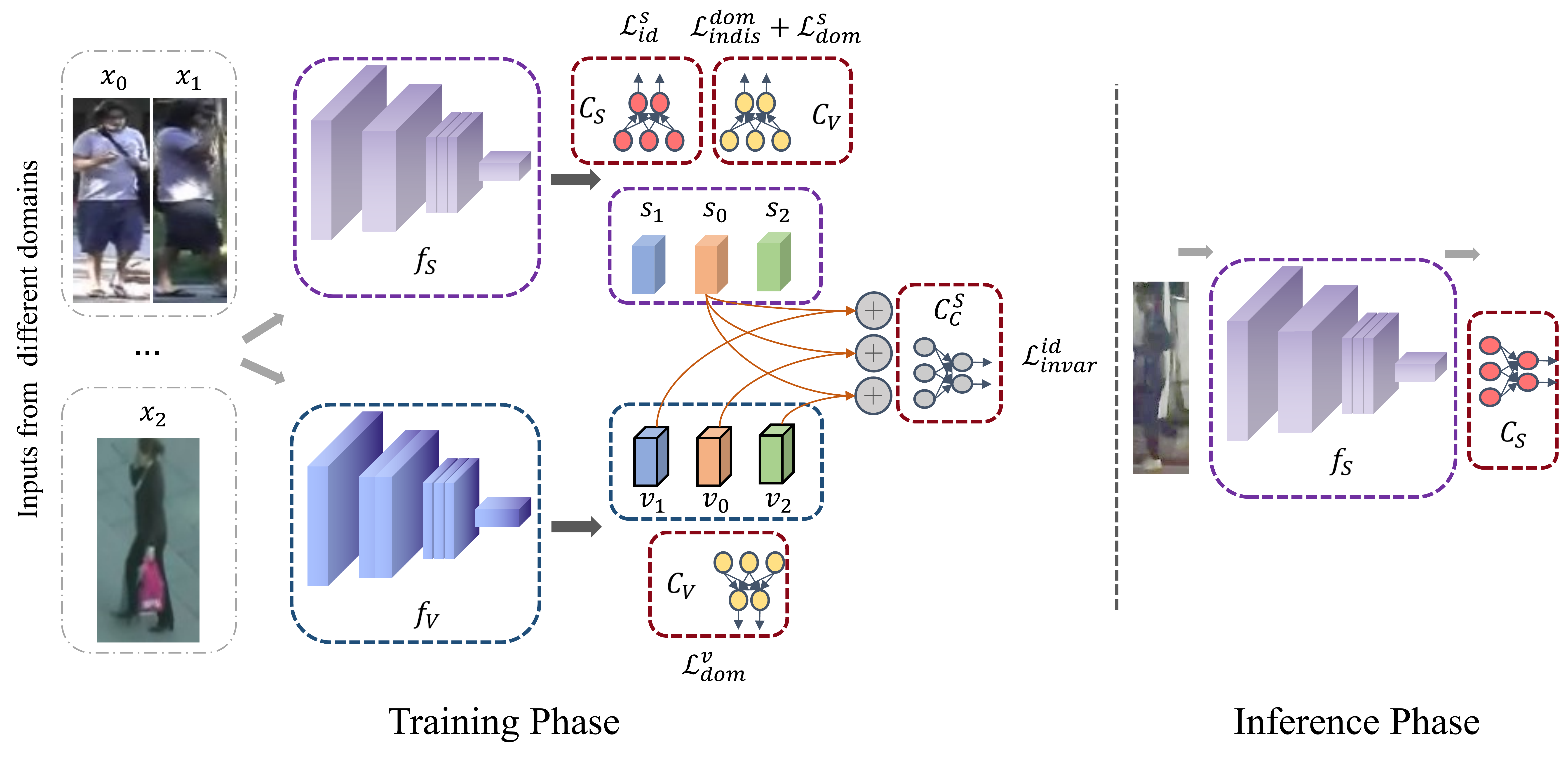}
  \end{center}
   \caption{\textbf{Schematic description of the proposed approach.} $x_0,x_1,x_2$: Three images from one mini-batch. $f_S, f_V$: encoders of identity-specific and domain-specific factors. $s,v$: identity-specific and domain-specific factors represented in latent space. $\mathcal{C}_S,\mathcal{C}_V, \mathcal{C}_C$: classifiers for identity-specific factors, domain-specific factors, and concatenated vectors. $\oplus$: concatenation of two latent vectors}
  \label{fig:model}
\end{figure*}
\subsection{The DIR-ReID Framework}\label{sec:framework}

\textbf{Notations and Problem Formulation.} For generalized person ReID, we have access to $G$ labeled datasets $\mathcal{D}=\{\mathcal{D}_g\}_{g=1}^{G}$. Each dataset $\mathcal{D}_g=\{({x}_i,y_i,d_i)\}_{i=1}^{N_g}$, where $N_g$ is the number of images in $\mathcal{D}_g$. The i-th data sample in $\mathcal{D}_g$ can be denoted as a triplet $(x_i, y_i, d_i)$, where $x_i,y_i,d_i$ denotes the image, identity label and the domain label respectively. In the training phase, we train a DG model using $N=\sum_{g=1}^GN_g$ aggregated image-label pairs from all source domains. In the testing phase, we perform a retrieval task on unseen target domains without additional model updates.

As analyzed in Section~\ref{sec:intervention}, the challenges are (i) how to get the representations of $S$ and $V$ from observation $X$. (ii) how to approximately marginalize over the domain-specific factors $V$. Here, we propose DIR-ReID to tackle these challenges. DIR-ReID consists of two blocks, as shown in \figurename~\ref{fig:model}.

(i) Multi-Domain Disentangled Adversarial Neural Network (MDDAN): MDDAN consists of two sub-blocks: (1) Identity adversarial learning block, which is a domain-agnostic, identity-aware encoder $f_S$ to obtain identity-specific factors. (2) Domain factors learning block, which is a domain-aware encoder $f_V$ to identify domain-specific factors.

(ii) Backdoor adjustment block, which approximates backdoor adjustment in Eq.\eqref{equ:intervention} based on the disentangled representation space $V$ and $S$.

 As shown in Figure \ref{fig:model}, the overall process includes feeding images $x_0, x_1, x_2$ randomly selected from the source domain $G$. into identity-specific and domain-specific encoders to obtain disentangled representations $s_0,s_1,s_2$ and $v_0,v_1,v_2$ via adversarial learning. Then backdoor adjustment is performed based on these representations to further train the encoders $f_S$ and $f_V$ for learning invariant representations.

\textbf{Identity Adversarial Learning Block.}
An identity-aware encoder $f_S$ is adopted to extract identity-specific factors. Then, a classifier  {$\mathcal{C}_S$} is used to identify the ID label for a given person image $x_i$. The cross-entropy loss with label smoothing~\cite{muller2019does} is calculated to train the encoder $f_S$, which is defined as:
\begin{equation}
  \mathcal{L}_{{id}}^{s}=-\sum_{i=1}^{N}\log{P\left(Y=y_i; \mathcal{C}_S\left(f_S\left(x_i\right)\right)\right)}
  \label{equ:idLoss}
\end{equation}
where $\theta_{S}$ is the parameters of $f_S$, $\mathcal{C}_S$ is the identity classifier. $y_i$ is the labeled person identity of $x_i$.

To exclude all the domain information from identity-specific factors, a domain classifier $\mathcal{C}_V$ is adopted for adversarial learning. One promising way is to use the gradient reversal layer (GRL) technique in DANN~\cite{ganin2016domain} to train the encoder $f_S$ and classifier $\mathcal{C}_V$ simultaneously, where a misclassification loss is adopted to force an image not to be classified in its true domain class. However, a desirable  disentangled representation should be ``indistinguishability'', rather than ``misclassification", which means classifying the input into all the domains equiprobably. Hence, we adopt a loss of maximum entropy (minimization of negative entropy~\cite{cai2019learning}) termed the domain-indistinguishability loss as follows.
\begin{equation}
  \mathcal{L}_{indis}^{dom}=- \mathcal{H}(D|S),
\end{equation}
where $\mathcal{H}$ is the entropy to measure the uncertainty of the predicted domain class given identity-specific factors, \ie,
\begin{equation}
  \begin{aligned}
   \mathcal{H}(D|S))=-&\sum_{i=1}^N P\left(D=d_i;\mathcal{C}_V\left(f_S\left(x_i\right)\right)\right)\\
   &\cdot\log P\left(D=d_i;\mathcal{C}_V\left(f_S\left(x_i\right)\right)\right),
  \end{aligned}
  \label{indisId}
\end{equation}
by which the extracted identity-specific features are required to reduce the domain information as less as possible.

As a result, the parameters $\theta_{S}$ of the identity-aware encoder $f_S$ are optimized by jointly minimizing the identity-classification loss and the domain-indistinguishability loss. The overall objective function of identity adversarial learning is as follows:
\begin{equation}
  \min_{\theta_{S}}\mathcal{L}^s_{id}+\lambda_1\mathcal{L}_{indis}^{dom},
\end{equation}
where $\lambda_1$ is a hyper-parameter to balance the trade-off between two losses. 

\textbf{Domain Factors Learning Block.} This block aims to extract domain-specific factors from person images $x_i$. 

The domain classification loss $\mathcal{L}_{dom}^v$ is defined as,
\begin{equation}
  \mathcal{L}^v_{dom}=-\sum_{i=1}^{N} \log P(D=d_i;\mathcal{C}_V(f_V(x_i))),
  \label{equ:domLoss}
\end{equation}
where $d_i$ is the domain label of image $x_i$.
The identity-indistinguishability loss $\mathcal{L}_{indis}^{id}$ is similar to Eq.\eqref{indisId}.

\textbf{Backdoor Adjustment Block.} With the disentangled representations learned from MDDAN, we can implement the backdoor adjustment. Given samples $\{x_i\}_{i=1}^N$, we first feed them to $f_S$ and $f_V$ to obtain $\{v_i\}_{i=1}^N$ and $\{s_i\}_{i=1}^N$. Then we follow two similar assumptions in~\cite{yue2020interventional}.

\noindent(i) $P(v_i)=1/|V|$, where we assume a uniform prior for the adjusted domain-specific factors.

\noindent(ii) $P(Y|S=s_i,V=v_k)=P(Y|{s_i}\oplus{v_k})$, where $\oplus$ denotes vector concatenation.

Based on the above assumption, the overall backdoor adjustment is as follows:

\begin{equation}
  P(Y|do(S=s_i))=\frac{1}{|V|}\sum_{k=1}^{|V|} P(Y|{s}\oplus{v_k})
\end{equation}

Here, to traverse all possible $V$, we propose 4 approximate implementations through random sampling over the feature space $V$.

\begin{itemize}
  \item $K$-Random. For each $x_i$, we randomly select $K$ different domain-specific factors $\{{v}_k\}^K_{k=1}$ where $k\neq i$.
  \item $K$-Hardest. For each $x_i$, we select $K$ domain-specific factors $\{{v}_k\}^K_{k=1}$ that are the most dissimilar to ${v_i}$.
  \item $K$-Mixup. For each $x_i$, we can create more domain-specific factors by mixup~\cite{zhang2018mixup,ye2020augmentation}. We generate $K$ mixed sample feature $\tilde{{v}}$ by interpolation of two randomly selected features (a pair $\langle {v}_{k},{v}_{k}'\rangle$), denoted by
  \begin{equation}
     \tilde{{v}} = \alpha {v}_{k} + (1-\alpha) {v}_{k}',
   \end{equation}
   where $\alpha\in [0, 1]$ controls the degree of interpolation and we empirically set $\alpha=0.5$ in our experiments.
  \item $K$-MixHard. For each $x_i$, we firstly select $K$ domain-specific factors $\{{v}_k\}_{k=1}^K$ least like ${v}_i$ and generate $K$ mixed features by randomly interpolating these factors.
\end{itemize}

Then we can approximate the backdoor adjustment by 
\begin{equation}
  P(Y|do(S=s_i))=\frac{1}{K}\sum_{k=1}^{K} P(Y|{s_i}\oplus{v_k})
\end{equation}

Taking into account the constant $1/K$, the final loss function is as follows.
\begin{equation}
  \mathcal{L}^{id}_{invar}=-\sum_{i=1}^{N}\sum_{k=1}^K \log P(Y=y_i;\mathcal{C}^S_C({s}_i\oplus{v_k})),
\end{equation}
where the $\mathcal{C}_C^S$ is a classifier for the concatenated feature vectors.

\subsection{Model Summary}
 Finally, given the parameters $\phi_S,\phi_V$ of the classifiers $\mathcal{C}_S,\mathcal{C}_V$, the total loss function is summarized as follows.
\begin{equation}
  \min_{\phi_S,\phi_V} \mathcal{L}^s_{id}+\mathcal{L}^s_{dom}+\mathcal{L}^v_{dom}.
  \label{equ:traincls}
\end{equation}
And the overall loss functions for the other components are defined as

\begin{equation}
  \begin{aligned}
   &\min_{\theta_{S},\phi_C^S} \mathcal{L}^s_{id}+\lambda_1\mathcal{L}^{dom}_{indis}+\lambda_2\mathcal{L}^{id}_{invar},\\
   &\min_{\theta_{V},\phi_C^V} \mathcal{L}^v_{dom}+\lambda_3\mathcal{L}^{id}_{invar},\\
  \end{aligned}
  \label{equ:trainother}
\end{equation}
where $\phi_C^S,\phi_C^V$ are parameters of classifiers $\mathcal{C}_C^S$ and $\mathcal{C}_C^V$ respectively. With the above components, each mini-batch training process is divided into two phases. In Phase I, the encoders $f_S$ and $f_V$ as well as the augmented data classifiers $\mathcal{C}_C^S$ and $\mathcal{C}_C^V$ are trained by Eq.\eqref{equ:trainother}, while $\mathcal{C}_S,\mathcal{C}_V$ are fixed. Then in Phase II, $\mathcal{C}_S,\mathcal{C}_V$ are trained by Eq.\eqref{equ:traincls}, while other components are fixed.

\subsection{Model Analysis}\label{sec:model_analysis}
In this section, we first give a theoretical analysis of MDDAN. Then, we state the differences between MDDAN and  other adversarial leaning-based methods. Finally, we discuss the proposed BA block in comparison with other causal representation methods.

\subsubsection{Theoretical characteristics of MDDAN.}

\begin{lemma}
  Let ${\tau}$ denote one of the source domains, ${s}^{j}$ are identity-specific factors of images from the $j$-th domain. Let $p({s}^j|\tau=i)$ be the class-conditional density function of the $j$-th domain given domain information $\tau=i$. It can be shown that carrying out MDDAN will lead to
  \begin{equation}
   p({s}^j|\tau=i)=p_j({s}^j), \forall \mathbf{s}^j,i=1,...,G.
  \end{equation}
  It indicates that in the latent space of identity-specific representation, the probability density will be invariant to different domains.
  Its class-conditional density function given any domains (\eg, $p({s}^j|\tau=i)$) is just equal to its prior density function value in its own domain (\eg, $p_j({s}^j))$, but not dependents on the domain variable $\tau$.\label{lemma:MDANN}
\end{lemma}
\begin{proof}
Referring to~\cite{Qi_2019_ICCV}, we can prove the Lemma~\ref{lemma:MDANN}. All the analyses are conducted in the shared identity-specific representation space. We first prove the following lemma.
\begin{lemma}
    When MDDAN is carried out to perform adversarial learning in Eq.\eqref{indisId}, given any identity-specific representation $\mathbf{s}^j$ of an image from domain $\tau=i$, the conditional probability of any $\tau=i$ given $\mathbf{s}^j$ will be equal to $1/G$, that is, for any $i=1,...,G$, we have $p(\tau=i|\mathbf{s}^j)=1/G$.
\end{lemma}~\label{lemma:UMS1}
\begin{proof}
    Here we slightly abuse the notation: omitting the domain classifier $\mathcal{C}_V$ and $ p(k|\mathbf{s}^j)$ is equivalent to $ p(\tau=k|\mathbf{s}^j)$. MDDAN is defined as follows.
    \begin{equation}
        \begin{split}
        &\max -\sum_{k=1}^G p(\tau=k|\mathbf{s}^j)\log p(\tau=k|\mathbf{s}^{j}), \\
        &\text{ s.t.}~\sum_{k=1}^G{ p(\tau=k|\mathbf{s}^j)}=1;  p(\tau=k|\mathbf{s}^j)\geq0, \forall k.
        \end{split}
        \end{equation}

    Let $\mathcal{H}(p_k)=p_{k}\log p_{k}$, where $p_k=p(\tau=k|\mathbf{s}^j)$. We simplify the above optimization problem as
    \begin{equation}
    \min\sum_{k=1}^G\mathcal{H}(p_k), \text{ s.t.}~\sum_{k=1}^G{p_{k}}=1;p_{k}\geq0,\forall k,
    \end{equation}

    where $\mathcal{H}(p_{k})$ is convex as we have $\mathcal{H}''(p_k)=\frac{1}{p_k\ln2}>0$.
    The sum of convex functions $\sum_{k=1}^G{\mathcal{H}(p_k)}$ is also a convex function. That is, this problem is a convex optimization problem. To prove the lemma, it is now equivalent to show that the minimum value of this convex optimization problem is obtained when $p_1=p_2=...=p_G=1/G$.
    
    We use \textit{augmented Lagrangian method} to solve the problem, where the Lagrange function is defined as
    \begin{equation}
        L(p,\lambda)=\sum_{k=1}^G{p_k\log p_k}+\lambda(\sum_{k=1}^G{p_k}-1).
    \end{equation}
    We take partial derivatives for each $p_k$ and get
    \begin{equation}
        \frac{\partial{L(p,\lambda)}}{\partial{p_k}}=\log{p_k}+\frac{1}{\ln2}+\lambda=0.
    \end{equation}
    Then we have $p_k=2^{-\lambda-1/\ln2}$. As $\sum_{k=1}^G{p_k}=1$, we have $G*2^{-\lambda-1/\ln2}=1$, and thus $p_k=1/G, i=1,2,...,G$.
    Since the local minimum of the convex function is the global minimum, when $p_1=p_2=...=p_G=1/G$, $\sum_{k=1}^G{\mathcal{H}(p_k)}$ achieves the minimum value, $\log \frac{1}{G}$.
    In other words, when MDDAN is carried out and achieves the maximum uncertainty, for all $i=1,...,G$, we have $p(\tau=i|\mathbf{s}^j)=1/G$.
\end{proof}
Now we are ready to prove Lemma 1. We can calculate any domain $\tau$'s conditional probability given $\mathbf{s}^j$, which is
\begin{equation}
    p(\tau=i|\mathbf{s}^j)=\frac{p(\mathbf{s}^j|\tau=i)p(\tau=i)}{p_j(\mathbf{s}^j)}, \forall \mathbf{s}^j; i=1,...,G,
\label{equ:equal}
\end{equation}
where $p(\mathbf{s}^j|\tau=i)$ denotes the conditional probability of $\mathbf{s}^j$ given domain information $\tau=i$, $p_j(\mathbf{s}^j)$ denotes the probability function of the identity-specific representation in its domain $\tau=j$, and $p(\tau=i)$ is the prior probability of domain classes $\tau=i$. Without generality, we set equal prior probability for each domain, namely $p(\tau=i)=1/G$. Further, from the Lemma~\ref{lemma:UMS1} we know that, optimizing MDDAN leads to $p(\tau=i|\mathbf{s}^j)=1/G$ for all $i=1,...,G$. Hence, \myref{equ:equal} becomes 

\begin{equation}
    \begin{split}
    &1/G=\frac{p(\mathbf{s}^j|\tau=i)1/G}{p_j(\mathbf{s}^j)}, \forall \mathbf{s}^j; i=1,...,G,\\
    \Rightarrow &p(\mathbf{s}^j|\tau=i)=p_j(\mathbf{s}^j), \forall \mathbf{s}^j; i=1,...,G,
    \end{split}
    \end{equation}
thus completing the proof.

\end{proof}
\begin{lemma}
  From the view of information theory, MDDAN is minimizing the mutual information between identity-specific factors $S$ and domain information variables $\tau$, namely $\min \mathcal{I}(S,\tau)$. 
\end{lemma}\label{app:info}
\begin{proof}

Minimizing the mutual information between the identity-specific factors $S$ and the domain information variables $\tau$ is defined as

\begin{equation}
    \begin{aligned}
         \min \mathcal{I}(S,\tau)&=\min \mathcal{H}(\tau)-\mathcal{H}(\tau|S)\\
         &=\min -\mathcal{H}(\tau|S)\\
         &=\max \mathcal{H}(\tau|S)
    \end{aligned}
\end{equation}

The second line is derived since the entropy of the domain distribution $\mathcal{H}(\tau)$ is not related to our optimization, which is only related to the statistics of the dataset. That is, our MDDAN is essentially minimizing mutual information between identity-specific factors and domain information.
\end{proof}

\subsubsection{Comparison with other adversarial learning-based methods}
Here we discuss similar studies DANN~\cite{ganin2016domain}, DDAN~\cite{chen2020dual} and CaNE~\cite{yuan2020calibrated}, which also use adversarial training to reduce domain divergence or nuisance divergence. There are three differences between our methods and them: (i) The \textbf{implementation strategies} are different. Given $n$ domains, DANN~\cite{ganin2016domain} needs $n$ binary classifiers to check which domain one image belongs to. DDAN~\cite{chen2020dual} selects one central domain, using one binary classifier to check if one image belongs to the central (1) or peripheral (0) domain. CaNE~\cite{yuan2020calibrated} implements adversarial training in nuisance attributes (camera ID and video timestamps), where a reweighted form of negative entropy is implemented to take into account class imbalance problems. We directly use entropy maximization with one multi-domain classifier, which extends the assumption of binary classification in DANN. (ii) The \textbf{roles} of adversarial learning are different. In our work, MDDAN is first performed to obtain an initial disentangled representation of the identity/domain-specific factors. Then, the MDDAN and BA blocks are implemented jointly to further optimize the disentangled representation. While other related work only uses adversarial training to enhance the invariance of learned representations to some nuisances.

 {
\subsubsection{Comparison with disentanglement-based methods}
Compared to \cite{cai2019learning}, our method has mainly three differences (1) \textbf{Setting.} \cite{cai2019learning} focus on domain adaptation, which has only two domains, i.e., the source domain and the target domain and a set of images in the target domain can be used for domain adaptation during the training phase. Ours are domain generalization, which has multiple domains and the target domain is unseen during training. (2) \textbf{Methodologies.} \cite{cai2019learning} has one decoder and uses the Evidence Lower Bound (ELBO) loss for training, which is not required for MDDAN. Furthermore, \cite{cai2019learning} uses traditional adversarial training + GRL to exclude domain-specific information, and we use the domain-indistinguishability loss. (3) We provide \textbf{mathematical explanations}. Firstly, we prove that the global optimal solution of MDDAN will lead to the independence between identity-specific and domain-specific factors (Lemma 3.1). Secondly, we prove that MDDAN is equivalent to minimizing mutual information between identity and domain factors (Lemma 3.3). 

There are also some other works on domain adaptation contributing to distinguishing domain-specific and domain-invariant features~\cite{chang2019domain}, however, \cite{chang2019domain} first estimates pseudo-labels for the examples in the target domain using the existing unsupervised domain adaptation algorithm and then learns the normalization layers for source and target domains separately. Though UDA-based methods improve the performance of cross-domain ReID to a certain extent, most of them require a large amount of unlabeled target data for model retraining, which is unrealistic for DG ReID. 
}

\subsubsection{Comparison with other causal representation learning methods} 
DIR-Reid has a similar causal graph with~\cite{wang2020visual,yang2020deconfounded,yue2020interventional}. ~\cite{wang2020visual,yang2020deconfounded,yue2020interventional} describe the causal mechanism in various visual tasks, while the elements in the SCM and the implementation details are entirely different. (i) \textit{Elements in the SCM.} The main differences are list in Table.~\ref{tab:scm}. As far as we know, it is the first attempt to use backdoor adjustment in the disentangled feature space (identity-specific and domain-specific feature spaces). (ii) \textit{Implementation details.} the implementation of ~\cite{yang2020deconfounded} is based on front-door adjustment. The implementation of~\cite{wang2020visual} is simple yet efficient: they concatenate the causal feature with the features of all the confounders and then use the concatenated feature to predict the label.~\cite{yue2020interventional} propose three kinds of implementations of backdoor adjustment, where the class-wise adjustment is most relevant to us. They concatenate the probabilistic combination of pretraining features of all classes with the causal feature to predict the label of the causal feature. The BA block in our work also implements backdoor adjustment by feature concatenation with a number of selected domain-specific feature vectors. Since it is untractable to get a well-defined confounder dictionary~\cite{wang2020visual} or pretraining features for all class~\cite{yue2020interventional}. It is natural to simply adopt the domain feature as the confounder dictionary\footnote{For each domain, the average of all the images' features will serve as the domain's feature. Namely, the confounder dictionary has $G$ items.}. However, it works poorly (the second row in Table~\ref{tab:ablation}), which indicates that it is indispensable to disentangle the identity-specific factors and domain-specific factors and use the proposed approximate implementations. 

\begin{table*}[]
\resizebox{\textwidth}{!}{
\centering
\begin{tabular}{ccccc}
\toprule
          &Field& Causal                & Effect           & Confounder         \\\hline
DIC~\cite{yang2020deconfounded}   &Image captioning    & Image                 & Caption          & Pretrained dataset \\
VC R-CNN~\cite{wang2020visual}&Image captioning, VQA& Object representation & Object label     & Context objects    \\
IFSL~\cite{yue2020interventional}&Few shot learning& Image & Label     & Pretrain knowledge    \\
BA (ours) &Domain generalization & Semantic factors      & Pedestrian label & Variation factors \\
\bottomrule
\end{tabular}}
\caption{Items in the SCM}\label{tab:scm}
\end{table*}

\section{Experiments}

\subsection{Datasets and Settings}\label{sec:dataset}
\begin{table}[]
\resizebox{0.48\textwidth}{!}{
  \centering
  \begin{tabular}{c|ccc}
  \toprule
  Collection & Dataset& IDs & Images \\ \hline
   \multirow{5}{*}{MS} & CUHK02& 1,816 & 7,264 \\ 
                     & CUHK03& 1,467 & 14,097 \\ 
                  & DukeMTMC-Re-Id & 1,812 & 36,411 \\ 
                     &Market-1501& 1,501 & 29,419 \\ 
                     &CUHK-SYSU& 11,934 & 34,547 \\ \bottomrule
  \end{tabular}}
 \caption{Training Datasets Statistics. All the images in these datasets, regardless of their original train/test splits, are used for model training.}\label{tab:traindata}
 \end{table}
 
 \begin{table}[]
 \resizebox{0.48\textwidth}{!}{
  \centering
  \begin{tabular}{c|c|c|c|c}
  \toprule
  \multirow{2}{*}{Dataset} & \multicolumn{2}{c|}{ Probe} & \multicolumn{2}{c}{ Gallery} \\\cline{2-5} 
   &      Pr. IDs&      Pr. Imgs&      Ga. IDs&  Ga. imgs    \\\hline
           PRID&      100& 100     &   649   &  649   \\
           GRID&      125& 125     &  1025    &  1,025   \\
           VIPeR&      316&  316  &  316   &  316  \\
           i-LIDS&      60&60     &     60 &  60   \\\bottomrule
 \end{tabular}}
 \caption{Testing Datasets statistics.}\label{tab:testdata}
 \end{table}
Following~\cite{Song_2019_CVPR,jia2019frustratingly}, we evaluate the DIR-ReID with multiple data sources (MS), where source domains cover five large-scale ReID datasets, including CUHK02~\cite{Li_2013_CVPR}, CUHK03 ~\cite{Li_2014_CVPR}, Market1501~\cite{Zheng_2015_ICCV}, DukeMTMC-ReID~\cite{Zheng_2017_ICCV}, CUHK-SYSU PersonSearch~\cite{xiao2016end}. The details of MS are summarized in Table~\ref{tab:traindata}. The unseen test domains are VIPeR~\cite{viper}, PRID~\cite{prid}, QMUL GRID~\cite{grid} and i-LIDS~\cite{ilids}. We follow the single-shot setting, where the number of probe/gallery images is summarized in Table~\ref{tab:testdata}. The average rank-k (R-k) accuracy and mean Average Precision (mAP) over $10$ random splits are reported based on the evaluation protocol. In this way, we simulate the real-world setting in which a ReID model is trained with all the public datasets and evaluate the generalization capability to unseen domains. Detailed evaluation protocols are as follows.

\textbf{GRID}~\cite{grid} contains $250$ probe images and $250$ true match images of the probes in the gallery. Also, there are a total of $775$ additional images that do not belong to any of the probes. We randomly remove $125$ probe images. The remaining $125$ probe images and $1025(775+250)$ images in the gallery are used for testing.

\textbf{i-LIDS}~\cite{ilids} has two versions, images and sequences. The former is used in our experiments. It involves $300$ different pairs of pedestrians observed in two disjoint camera views $1$ and $2$ in open public spaces. We randomly select $60$ pedestrian pairs, two images per pair are randomly selected as probe image and gallery image, respectively.

\textbf{PRID2011}~\cite{prid} has single shot and multishot versions. We use the former in our experiments. The single-shot version has two camera views $A$ and $B$, which capture $385$ and $749$ pedestrians, respectively. Only $200$ pedestrians appear in both views.  During the evaluation, $100$ randomly identities presented in both views are selected, the remaining $100$ identities in view $A$ constitute the probe set, and the remaining $649$ identities in view $B$ constitute the gallery set.

\textbf{VIPeR}~\cite{viper} contains $632$ pairs of pedestrian images. Each pair contains two images of the same individual seen from different camera views $1$ and $2$.  Each pair of images was taken from an arbitrary viewpoint under varying illumination conditions. To compare to other methods, we randomly select half of these identities from camera view $1$ as probe images and their matched images in view $2$ as gallery images.

\subsection{Implementation Details}

Following previous generalizable person ReID methods, we use MobileNetV2~\cite{sandler2018mobilenetv2} as the domain-specific encoder $f_V$ and use MobileNetV2 with IN layer~\cite{lin2018multi} as identity-specific encoder $f_S$. Our classifiers $\mathcal{C}_S,\mathcal{C}_V,\mathcal{C}_C^S,\mathcal{C}_C^V$ are simply composed of a single fully-connected layer. Images are resized to $256\times 128$ and the training batch size is set to $128$. Random cropping, random flipping, and color jitter are applied as data augmentations. The label smoothing parameter is $0.1$. SGD is used to train all components from scratch with a learning rate of $0.02$ and a momentum of $0.9$. The training process includes $150$ epochs and the learning rate is divided by 10 after 100 epochs. At test time, DIR-ReID only involves identity-specific encoder $f_S$, which is of a comparable network size to most ReID
methods. The tradeoff weights are set to $\lambda_2=0.1$ and $\lambda_1=\lambda_3=\lambda_4=1$ empirically.
\subsection{Comparisons Against State-of-the-art}
\begin{table*}[]
\resizebox{\textwidth}{!}{
  \centering
  \begin{tabular}{c|c|c|cccc|cccc|cccc|cccc}
  \toprule
  \multirow{2}{*}{Methods} & \multirow{2}{*}{Type} & \multirow{2}{*}{Source} & \multicolumn{4}{c|}{VIPeR (V)} & \multicolumn{4}{c|}{PRID (P)} & \multicolumn{4}{c|}{GRID (G)} & \multicolumn{4}{c}{i-LIDS (I)} \\ \cdashline{4-19} 
              & & & R-1 & R-5 & R-10 & $m$AP&R-1& R-5 & R-10& $m$AP& R-1& R-5 & R-10& $m$AP & R-1& R-5 & R-10& $m$AP \\ \hline
              DeepRank~\cite{chen2016deep}&S& Target&38.4 & 69.2 &81.3& -&-&-& -& - &-&-&-&-& -&-& -& -    \\ 
              DNS~\cite{Zhang_2016_CVPR}&S& Target&42.3& 71.5&82.9 &-&29.8& 52.9 &66.0 & -& - &-&-&-&-& -&-& -    \\ 
              MTDnet~\cite{chen2017multi}&S&Target&47.5& 73.1& 82.6 &-&32.0& 51.0 &62.0 & -&-&-& -&-&58.4&80.4&87.3&-  \\ 
              JLML~\cite{Zhang_2016_CVPR}&S& Target&50.2& {\color{red}74.2}& {\color{red}84.3} &-&-&-&-&-&37.5&{61.4} &69.4 &-&-& -&-& -  \\ 
              SSM~\cite{Bai_2017_CVPR}&S& Target&53.7& -& {\color{blue}91.5} &-&-&-&-&-&27.2& - &61.2 &-&-& -&-& -  \\ 
              SpindleNet~\cite{Zhao_2017_CVPR}&S& Target&{\color{red}58.3}& {74.1}& 83.2&- &67.0&{\color{blue}89.0}&{\color{red}89.0}&-&-&-&-&-&66.3&86.6& 91.8& -  \\ \hline
              AugMining~\cite{tamura2019augmented}&DG&MS&49.8 &70.8&77.0&-&34.3&56.2&65.7&-&{\color{red}46.6}&{\color{blue}67.5}&{\color{red}76.1}&-&76.3&{\color{red}93.0}&{\color{red}95.3}&- \\
              DIMN~\cite{Song_2019_CVPR}&DG&MS&51.2&70.2&76.0&60.1&39.2&67.0&76.7&52.0&29.3&53.3&65.8&41.1&70.2&{89.7}&{94.5}&78.4 \\ 
              DualNorm~\cite{jia2019frustratingly}&DG&MS&{53.9}&62.5&75.3&58.0&{60.4}&73.6&84.8&{64.9}&41.4&47.4&64.7&{45.7}&{74.8}&82.0&91.5&{78.5}\\ 
            
             SNR~\cite{jin2020style}&DG&MS&52.9&-&-&{\color{red}61.3}&52.1&-&-&66.5&40.2&-&-&47.7&{\color{blue}84.1}&-&-&{\color{blue}89.9}\\
              DDAN~\cite{chen2020dual}&DG&MS&{56.5}&65.6&76.3&60.8&{\color{red}62.9}&74.2&85.3&{\color{red}67.5}&46.2&55.4&68.0&50.9&{78.0}&85.7&93.2&81.2\\ 
              \rowcolor{gray!40} DIR-ReID&DG&MS&\textbf{\color{blue}58.5}&{\color{blue}76.9}&83.3&\textbf{\color{blue}67.0}&\textbf{\color{blue}69.7}&{\color{red}85.8}&{\color{blue}91.0}&\textbf{\color{blue}77.1}&\textbf{\color{blue}48.2}&{\color{red}67.1}&{\color{blue}76.3}&\textbf{\color{blue}57.6}&{\color{red}79.0}&{\color{blue}94.8}&{\color{red}97.2}&{\color{red}83.4}\\ \bottomrule
  \end{tabular}}
  \caption{ 
  Comparisons against state-of-the-art methods. 'S': single domain, 'DG': domain generalization, 'M': Market1501, 'D': DukeMTMC-ReID, 'comb': the combination of VIPeR, PRID, CUHK01, i-LIDS, and CAVIAR datasets. 'C3': CUHK03, '-': no report. $1^{st}$ and $2^{ed}$ highest accuracy are indicated by \textbf{\color{blue}blue} and {\color{red}red} color.}\label{tab:sota}
  \end{table*}

\begin{table}[]
\resizebox{0.48\textwidth}{!}{
\begin{tabular}{c|cccccccc}
\toprule
\multirow{3}{*}{Method} & \multicolumn{8}{c}{Cross-domain Re-ID (single-source DG)}                                   \\\cdashline{2-9}
            & \multicolumn{4}{c}{Market-Duke}                & \multicolumn{4}{c}{Duke-Market}                \\\cdashline{2-4}\cdashline{4-9}
            & R-1      & R-5      & R-10     & $m$AP      & R-1      & R-5      & R-10     & $m$AP      \\\hline
IBN-Net~\cite{pan2018two}         & 43.7     & 59.1     & 65.2     & 24.3     & 50.7     & 69.1     & 76.3     & 23.5     \\
OSNet~\cite{zhou2021learning}          & 44.7     & 59.6     & 65.4     & 25,9     & 52.2     & 67.5     & 74.7     & 24.0     \\
OSNet-IBN~\cite{zhou2021learning}        & 47.9     & 62.7     & 68.2     & 27.6     & 57.8     & 74.0     & 79.5     & 27.4     \\
CrossGrad~\cite{shankar2018generalizing}        & 48.5     & 63.5     & 69.5     & 27.1     & 56.7     & 73.5     & 79.5     & 26.3     \\
QAConv~\cite{liao2019interpretable}         & 48.8     & -       & -       & 28.7     & 58.6     & -       & -       & 27.6     \\
L2A-OT~\cite{zhou2020learning}         & 50.1     & 64.5     & 70.1     & 29.2     & 63.8     & 80.2     & 84.6     & 30.2     \\
OSNet-AIN~\cite{zhou2021learning}        & 52.4     & 66.1     & 71.2     & 30.5     & 61.0     & 77.0     & 82.5     & 30.6     \\
SNR~\cite{jin2020style}           & {\color{blue}{55.1}} & -       & -       & {\color{blue}33.6} & {\color{red}66.7}     & -       & -       & {\color{red}33.9}     \\
DIR-ReID        & {\color{red}54.5}     & {\color{blue}66.8} & {\color{blue}72.5} & {\color{red}33.0}     & {\color{blue}68.2} & {\color{blue}{80.7}} & {\color{blue}86.0} & {\color{blue}35.2}\\\bottomrule

\end{tabular}}
\caption{performance (\%) comparison with the state-of-the-arts on the cross-domain ReID problem. $1^{st}$ and $2^{ed}$ highest accuracy are indicated by \textbf{\color{blue}blue} and {\color{red}red} color.}\label{tab:uda}
\end{table}
\textbf{Comparison with single domain methods.} 
Many supervised methods report high performance on large-scale benchmarks, but their performance is still poor on small-scale. We select $6$ representative models (labeled as `S' in Table~\ref{tab:sota}) in comparisons, which are trained with the data splits in the target datasets. Although data from the target domain are inaccessible for DIR-ReID, it achieves competitive or better performance on all four benchmarks, which indicates that sufficient source data and our model based on domain invariance learning can alleviate the need for data from the target domain.

\textbf{Comparison with DG methods.}
Then, we compare DIR-ReID with existing methods on generalizable person ReID. As far as we know, there are a few publications focusing on person ReID generalization problem, including DIMN~\cite{Song_2019_CVPR}, DualNorm~\cite{jia2019frustratingly}, ~\cite{tamura2019augmented} and DDAN~\cite{chen2020dual}. From the third row in Table~\ref{tab:sota}, DIR-ReID has achieved the best performance in terms of mAP against other SOTA DG-ReID methods. Although our method falls behind others on the i-LIDS and the GRID datasets in terms of Rank-5 and Rank-10, the DIR-ReID obtains the best Rank-1 performance on three of four datasets. Interestingly, methods such as SNR~\cite{jin2020style} and AugMining~\cite{tamura2019augmented} perform very well in i-LIDS, while having low performance in other datasets, suggesting the unstable generalization abilities of their models. To further measure the generalization ability, we adopt the worst-domain accuracy (WDA) proposed in~\cite{sagawa2019distributionally} for a comparison. From Table~\ref{tab:sota}, we can find that our DIR-ReID achieves the highest WDA value with $47.8\%$ rank-1 accuracy, which demonstrates the superior generalization ability of our model.


\textbf{Comparison with cross-domain Re-ID methods.}

To further evaluate the generalization ability of our approach, we also perform cross-domain ReID tests with two large-scale datasets, \ie, Market1501 and DukeMTMC. The experimental results are presented in Table.~\ref{tab:uda}(``Market1501$\rightarrow$DukeMTMC" indicates that Market-1501 is a labeled source domain and DukeMTMC-ReID is an unseen target domain). It is different from the settings in UDA methods, all models in our comparisons only use the source data for training without any model adaptations in the target domain. The setting of cross-domain Re-ID is challenging for us because there is only one source domain. Thus we consider each camera view as a single domain for training the MDDAN block. As the camera views in the same dataset may share similar imaging characteristics, \eg, background environments, and resolutions, the variations of domain-specific factors may be smaller within a single dataset than that of multiple data sources. Despite this, Table.~\ref{tab:uda} shows that our DIR-ReID framework achieved comparable performance on both settings. It indicates that, even with small domain variations, DIR-ReID can improve the generalization capability.

It is noted that DIR-ReID still has a large margin with current UDA methods (\eg~MMT~\cite{ge2020mutual} achieves more than $75\%$ rank 1 accuracy in the ``Market-to-Duke" dataset setting, which is much superior to current DG methods). DIR-ReID  does not need any model adaptations with the data of the target domain, which significantly reduces the costs of large-scale data collections in practical deployments of ReID models.

\subsection{Ablation Studies}
There are two main components in the proposed DIR-ReID: the MDDAN block and the BA block. Here, we first analyze the effectiveness of each block respectively, then demonstrate their contributions to the final performance of the whole DIR-ReID model.

\textbf{Effectiveness of MDDAN.} The superiority of MDDAN is verified by comparisons with the dual DANN~\cite{ganin2016domain} block. The latter means inserting the GRL layers between $f_S,C_V$ and $f_V,C_S$. Simultaneously, the dual DANN block replaces the maximum entropy loss in MDDAN with the maximum misclassification loss. The results are shown in Table~\ref{tab:ablation}. The dual DANN method only has an improvement with average $1.1$ points over 4 test datasets in comparison with the baseline, while our MDDAN block leads to significant improvements.
\begin{figure}[h]
  \begin{center}
   \includegraphics[width=1.0\linewidth,scale=1.0]{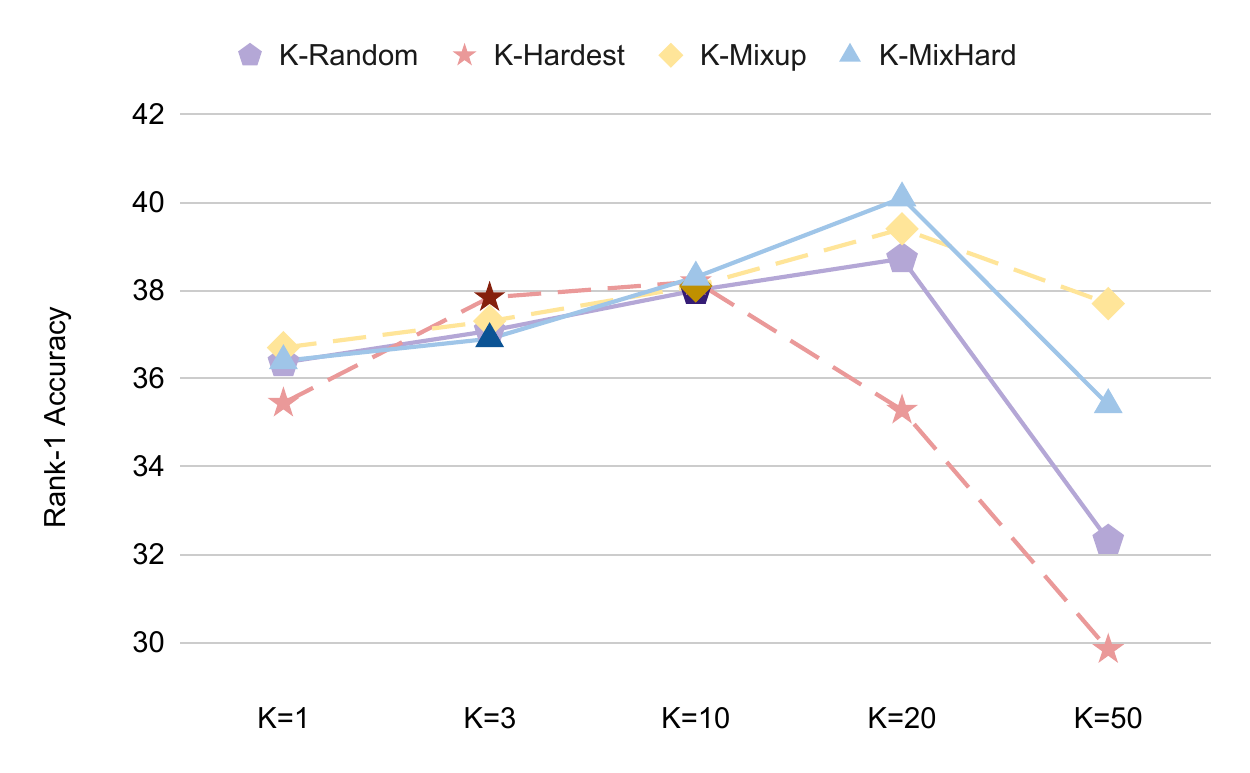}
  \end{center}
   \caption{Ablation study on BA block. The metric is the rank-1 accuracy on the GRID dataset. Considering the expensive cost of training with five datasets, all the models here are trained on three datasets, \ie~Market-1501, CUHK02 and CUHK03.}
  \label{fig:ablation-K1}
\end{figure}

\textbf{Effectiveness of BA block.} Firstly we compare the proposed BA block with Confounder Dict, which imitates similar backdoor adjustment-based methods~\cite{wang2020visual,yue2020interventional}. The difference between Confounder Dict and DIR-ReID is the source of variation factors. Specifically, Confounder Dict uses a pretrained MobileNetV2 (pretrained in ImageNet) to extract all the features of the data points. The average feature of the images in domain $g$ will serve as its confounder feature $v_g$ and $\{v_g\}_{g=1}^G$ will serve as domain-specific factors. Then BA block can be executed by concatenating the identity-specific factors extract by $f_S$ and the pretrained domain-specific factors. The results are shown in Table~\ref{tab:ablation}. The confounder Dict is inferior to DIR-ReID and we believe there are two reasons: (i) The pretrained MobileNetV2 cannot extract the proper factors in surveillance scenarios \eg~backgrounds and illumination conditions. (ii) The number of confounder features $G$ is small and cannot cover the variation space at all. On the contrary, the DIR-ReID trains a $f_V$ to extract variation factors for each person image to obtain specific domain factors. Hence it is more potential to approximate the variation space and finish the backdoor adjustment. We also compare four different implementations of BA and the results are shown in~\figurename~\ref{fig:ablation-K1}. $K$-Random is the simplest method while it works well, attaining $38.72\%$ rank 1 accuracy when $K=20$. $K$-MixHard outperforms other methods and attains $40.16\%$ rank 1 accuracy, which verified the importance of mixup for data augmentation. However, as we increase the value of $K$, the performance will not be improved. 


\textbf{Ablation study of different blocks.}
\begin{table}[]
\resizebox{0.5\textwidth}{!}{
  \centering
  \begin{tabular}{c|c|c|c|c}
  \toprule
              & PRID & VIPeR & i-LIDS &GRID \\\hline
  Baseline (DualNorm~\cite{jia2019frustratingly})  &  38.9  &  54.0  & 64.3 & 34.4 \\\hline
  w/ Dual DANN~\cite{ganin2016domain}      &  41.0 &  53.3  &  66.2 &35.4 \\
  Improvements$\uparrow$      & 2.1 &  -0.7  &  1.9 &1.0\\\hline
    w/ Confounder Dict~\cite{wang2020visual,yue2020interventional}     &  39.4 &  54.7  &  65.8 &35.9 \\
  Improvements$\uparrow$      & 0.5 &  0.7  &  1.5 &1.5\\\hline
  w/ MDDAN       & 42.7 &  56.5  &  65.7 &36.6\\
  Improvements$\uparrow$      & 3.8 &  2.5  &  1.4 &2.2\\\hline
  \rowcolor{mygray}
  w/ BA  &  43.8 &  61.4 & 68.2 &40.2 \\
  Improvements$\uparrow$  & 4.9 &  7.4  &  3.9 &5.8 \\\hline
  w/ Triplet Loss  &  44.6 &  62.7 & 68.9 &41.4 \\
  Improvements$\uparrow$  & 5.7 &  8.7  &  4.6 &7.0 \\\bottomrule
  \end{tabular}}
  \caption{Ablation studies on three different blocks: dual DANN block, Confounder dictionary block, the proposed MDDAN block and the BA block. The metric is rank-1 accuracy. Models here are all trained on three datasets, including Market-1501, CUHK02 and CUHK03. The improvements denote the difference from the baseline.}
  \label{tab:ablation}
  \end{table}
  To evaluate the contribution of each component, we gradually add the MDDAN and the BA to the baseline, and the overall ablation studies are reported in Table~\ref{tab:ablation}. The MDDAN improves the rank-1 accuracy from $34.4\%$ to $36.6\%$ in the GRID dataset. The results in PRID, VIPeR, and i-LIDS datasets are consistently improved, which validates that the MDDAN removes some of the domain-specific information from the identity-specific representations and yields consistent generalization performance improvements. The BA provides greater improvement gains on three test datasets. It validates that BA can exclude domain-specific information efficiently. {Besides, We conduct an ablation study on the triplet loss, which is also shown able to boost performance, indicating that the metric learning method is orthogonal to our DIR-ReID framework (Last line in Table~\ref{tab:ablation}).}

 {
\textbf{Ablation studies of every loss function.} A thorough ablation is shown in Table~\ref{tab:ablation_losses}. Incorporated with $\mathcal{L}_{indis}^{dom}$ for the encoder $f_S$, the performance of the baseline where only $\mathcal{L}_{id}^s$ for $f_S$ exists can be improved from $47.98\%$ to $50.38\%$. If we implement BA block only with $\mathcal{C}_C^S,f_S,f_V$ trained by $\mathcal{L}^{id}_{invar}$ or $\mathcal{L}^{id}_{invar}+\mathcal{L}^{s}_{id}$, namely no constraint on $f_V$, the performance is similar to the baseline because the feature space is not disentangled well and the backdoor adjustment cannot attain better performance. Once $f_V$ is further trained by $\mathcal{L}_{dom}^v$, namely, $f_V$ is constrained to contain domain information, the performance gets better. Finally, the $\mathcal{L}_{indis}^{dom}$ is incorporated and $f_S$ is forced to remove domain information, the proposed DIR-ReID attains the best performance in such a disentangled feature space.}

\begin{table}[]
\centering
\resizebox{0.48\textwidth}{!}{
\begin{tabular}{@{}cccccc@{}}
\toprule
                                          & PRID & VIPeR & i-LIDS & GRID & Avg   \\ \midrule
Baseline (only $\mathcal{L}_{id}^s$)                                                                                                                                                                                & 38.9 & 54.0  & 64.3   & 34.4 & 47.98 \\
MDDAN ($\mathcal{L}_{id}^s+\mathcal{L}_{indis}^{dom}$)                                                                                                                                                                          & 42.7 & 56.5  & 65.7   & 36.6 & 50.38 \\
$\mathcal{L}^{id}_{invar}$                                                                                                                                                                                                   & 39.1 & 53.7  & 63.2   & 35.2 & 47.80 \\
$\mathcal{L}^{id}_{invar}+\mathcal{L}_{id}^s$                                                                                                                                  & 39.1 & 54.2  & 64     & 34.8 & 48.03 \\
$\mathcal{L}^{id}_{invar}+\mathcal{L}_{id}^s+\mathcal{L}_{dom}^v$                                                                 & 40.0   & 58.2  & 66.7   & 39.6 & 51.13 \\
$\mathcal{L}^{id}_{invar}+\mathcal{L}_{id}^s+\mathcal{L}_{indis}^{dom}+\mathcal{L}_{dom}^v$ & 43.8 & 61.4  & 68.2   & 40.2 & 53.40 \\ \bottomrule
\end{tabular}}
\label{tab:ablation_losses}
\caption{ {Ablation studies of every loss function. The metric is rank-1 accuracy. Models here are all trained on three datasets, including Market-1501, CUHK02, and CUHK03. The improvements denote the difference from the baseline.}}
\end{table}

We also conduct additional ablation studies on rotated MNIST. The dataset and results are as follows:

\subsection{Ablation studies on Rotated MNIST.}
Since the ReID datasets are collected from real surveillance scenarios, complex data variations hinder us from analyzing the characteristics of the proposed model for feature disentanglement. Thus, we performed additional studies on a controlled simple dataset, \ie, the rotated MNIST.
\begin{figure*}[h]
    \centering
    \subfigure[]{
      \begin{minipage}[t]{0.45\linewidth}
      \centering
      \includegraphics[scale=1.0]{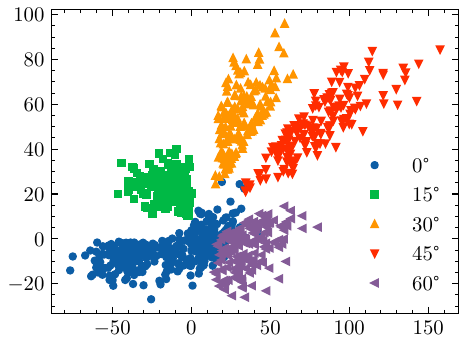}\label{fig:mnist1}
       \end{minipage}}
       \hfill
       \subfigure[]{
       \begin{minipage}[t]{0.45\linewidth}
       \centering
      \includegraphics[scale=1.0]{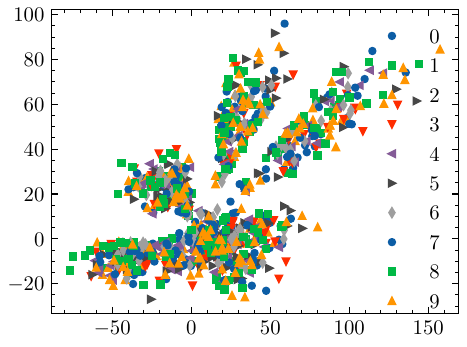}\label{fig:mnist2}
       \end{minipage}}
         \hfill
         \subfigure[]{
         \begin{minipage}[t]{0.45\linewidth}
         \centering
        \includegraphics[scale=1.0]{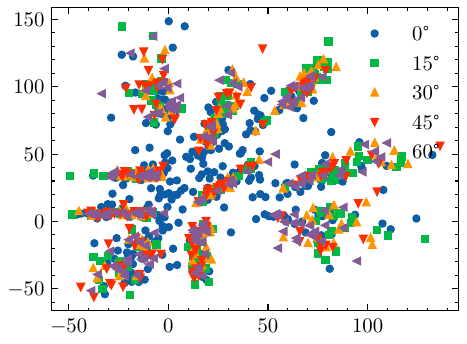}\label{fig:mnist3}
         \end{minipage}}
         \hfill
         \subfigure[]{
          \begin{minipage}[t]{0.45\linewidth}
          \centering
         \includegraphics[scale=1.0]{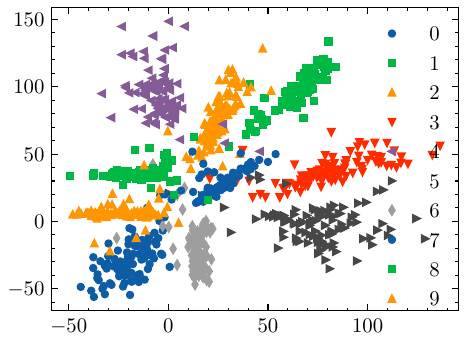}\label{fig:mnist4}
          \end{minipage}}
       \caption{2D visualizations of all two latent subspaces. (a,b): ${v}$ encoded by $f_V$. (c,d): ${s}$ encoded by $f_S$. (a,c) are colored according to their domains and (b,d) are colored according to their classes.}\label{fig:mnist}
 \end{figure*}
\subsubsection{Dataset and Setting}
To verify the capability of the DIR-ReID model to disentangle $S$ and $V$, we first construct rotated MNIST datasets as follows~\cite{Ghifary_2015_ICCV}. $100$ images per class ($10$ classes totally) are randomly sampled from the MNIST training dataset, which is denoted by $\mathcal{M}_{0^\circ}$. We then rotated the images in $\mathcal{M}_{0^\circ}$ by $15,30,45,60$ and $75$ degrees, creating five additional domains. Models are trained in $\{\mathcal{M}_{0^\circ},...,\mathcal{M}_{60^\circ}\}$ and tested on $\mathcal{M}_{75^\circ}$. To plot the latent subspaces directly without applying dimensionality reduction, we restrict the size of latent spaces for $S,V$ to 2 dimensions. In experiments, we train the MDDAN block with an additional reconstruction loss, which is enough to attain an encouraging result. 
\subsubsection{Architecture and Implementation Details}
The architectures of the encoders, classifiers are shown in Table~\ref{tab:arc_enc}, Table~\ref{tab:arc_cls} respectively. Our model and the baseline model Dual DANN are trained for 500 epochs and the batch size is set to 100. Adam optimizer is used to train all components from scratch with a learning rate of 0.001. We also use warm-up to linearly increase the learning rate from 0 to 0.001 during the first 100 epochs of training. 


\begin{table}
\centering
\resizebox{0.4\textwidth}{!}{
    \begin{tabular}{c|c}
        \toprule
        block & details\\ \hline
        1 & Conv2d(32, 5), BatchNorm2d, ReLU\\
        2 & MaxPool2d(2, 2)\\
        3 & Conv2d(64, 5), BatchNorm2d, ReLU \\
        4 & MaxPool2d(2, 2)\\
        5 & Linear(2)\\\bottomrule
    \end{tabular}}
    \caption{Architecture for encoders $f_S, f_V$ used in the ablation studies on rotated MINIST dataset. The parameters for Conv2d are output channels and kernel size. The parameters for MaxPool2d are kernel size and stride. The parameter for Linear is output features.}\label{tab:arc_enc}
\end{table}

\begin{table}
\centering
\resizebox{0.25\textwidth}{!}{
    
    \begin{tabular}{c|c}
        \toprule
        block & details\\ \hline
        For $\mathcal{C}_V$ &ReLU,Linear($5$ ) \\
        For $\mathcal{C}_S$ &ReLU,Linear($10$) \\ 
        For $\mathcal{C}_C^S$ &ReLU,Linear($5$ ) \\
        For $\mathcal{C}_C^v$ &ReLU,Linear($5$ ) \\\bottomrule
    \end{tabular}}
    \caption{In the ablation studies on rotated MINIST dataset: Architecture of classifiers for ID-specific factors, Domain-Specific factors, and
    concatenated vectors. The parameter for Linear is output features.}\label{tab:arc_cls}
\end{table}

\subsubsection{Additional Experimental Results}
 
\begin{table}[]
    \centering
\resizebox{0.4\textwidth}{!}{

    \begin{tabular}{c|c}
    \toprule
    & Test Accuracy on $\mathcal{M}_{75^\circ}$ \\\hline
    Baseline   & $46.4$    \\\hline
    w/ Dual DANN~\cite{ganin2016domain}           & $53.5$      \\
    Improvements$\uparrow$            & $7.1$    \\\toprule
    w/  MDDAN           & $58.2$      \\
    Improvements$\uparrow$            & $11.8$    \\\hline
    w/ BA & $61.9$    \\
    Improvements$\uparrow$  & $15.5$   \\\toprule
    \end{tabular}}
    \caption{Ablation study of three different blocks: dual GRL block, our MDDAN block and the BA block. The reported validation metric is the accuracy of the $\mathcal{M}_{75^\circ}$ dataset. The baseline is only using the encoder $f_S$ and classifier $\mathcal{C}_S$.}
    \label{tab:ablation-mnist}
    \end{table}

    \begin{figure}[h]
        \begin{center}
           \includegraphics[width=1.0\linewidth,scale=1.0]{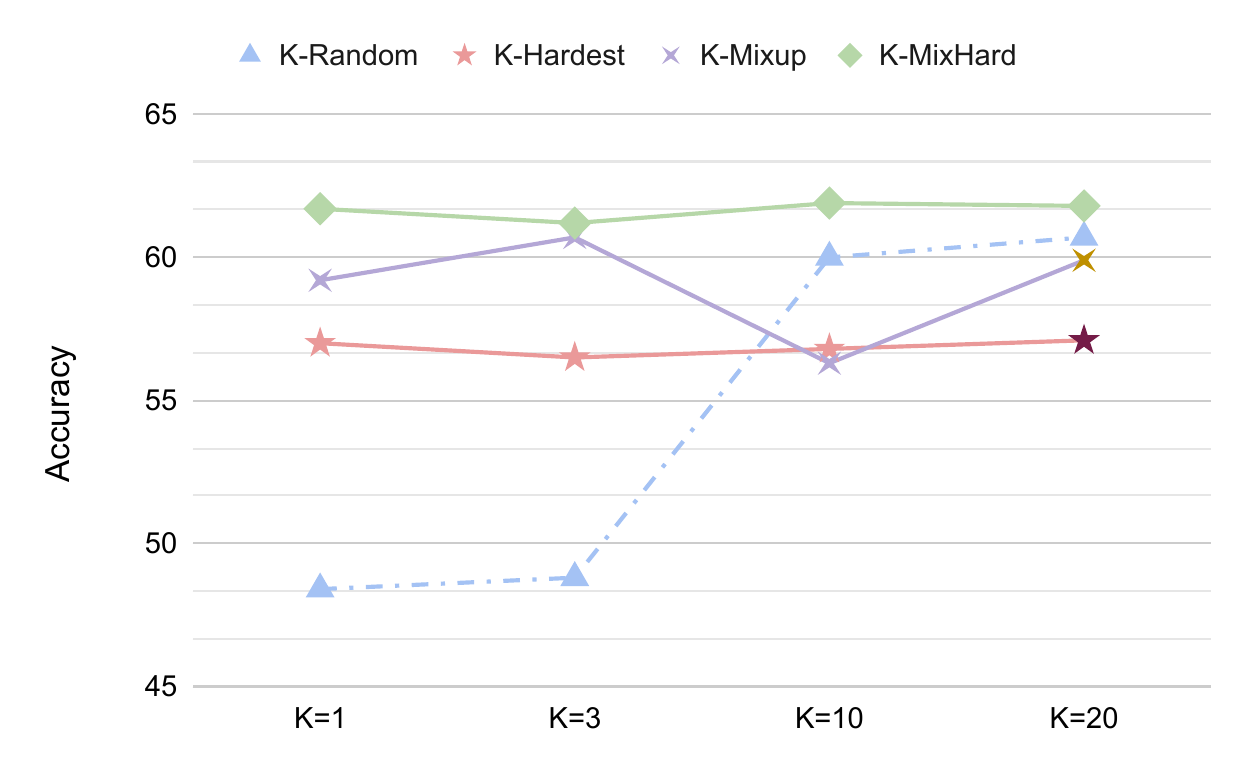}
        \end{center}
           \caption{Ablation study of backdoor adjustment methods. The reported validation metric is the test accuracy of the $\mathcal{M}_{70\circ}$ dataset.}
        \label{fig:ablation-K}
     \end{figure}
\textbf{Analysis of MDDAN.} As shown in Table~\ref{tab:ablation}, directly applying the dual GRL block benefits the generalization ability, while the proposed MDDAN improve the test accuracy on $\mathcal{M}_{75^\circ}$ dataset by an even more large margin, which is $11.8$ points.

\textbf{Analysis of methods for backdoor adjustment.} The comparison results are shown in~\figurename~\ref{fig:ablation-K}. Similar to BA for Re-ID, here $K$-MixHard attains the most superior performance, which is $61.9\%$ test accuracy.

\textbf{Ablation study of different blocks.} By adding the multi-domain disentangled block and the backdoor adjustment block successively, we improve the generalization accuracy from $46.4\%$ to $58.2\%$ and $61.9\%$ respectively (Table~\ref{tab:ablation-mnist}), showing the effectiveness of the proposed model again.

\textbf{Visualization analysis.} 
The disentanglement results are visualized in~\figurename~\ref{fig:mnist}.
We can find a correlation between the rotation angle (domain labels) and the learned domain-specific features ${V}$ in~\figurename~\ref{fig:mnist1}, five domains are clustered into five distinct clusters, while in~\figurename~\ref{fig:mnist2} no clustering is visible, which denotes the very weak correlations between $V$ and class labels. By contrast, in~\figurename~\ref{fig:mnist3} no clustering is visible according to the domain labels of rotation angles. But~\figurename~\ref{fig:mnist4} shows ten distinct clusters, where each cluster corresponds to a digit class. From these  qualitative results, we conclude that the MDDAN is able to disentangle the information contained in the rotated MNIST dataset, where the learned latent subspaces indeed encode the domain (rotation angles) information and identity (digit classes) information respectively.

\section{Conclusions and Future Work}

We propose a novel generalizable person ReID approach based on disentanglement and backdoor adjustment from a causal invariance learning framework. Specifically, a MDDAN block is proposed to disentangle identity-specific and domain-specific factors from multi-source ReID training data. We then propose a BA block to learn the interventional distribution and reduce the confounding effects via backdoor adjustment. The comprehensive experimental results show that DIR-ReID achieves state-of-the-art performance.

In future, we can improve the model performance with other regularization techniques. One promising way is to generate realistic images from the latent disentangled representations. The augmented feature vectors are guided by a reconstruction loss, which will further improve the disentanglement of identity-specific and domain-specific factors. These generated images can also be used for augmenting the training set. Besides, we will seek other methods for better disentanglement performance such as replacing the multi-domain adversarial learning with mutual information minimization~\cite{belghazi2018mine} or $f$-divergence maximization~\cite{nowozin2016f}.

\section{Acknowledgement}
This work is funded by the National Natural Science Foundation of China (Grant No. 62106260), China Postdoctoral Science Foundation (Grant No. 2020M680751), and National Natural Science Foundation of China (62236010, 61721004, and U1803261).

\appendices
\section{Derivation of the Interventional Distribution $P(Y|do(S))$ for the Proposed Causal Graph}\label{proof:intervention}
The following proof is similar to~\cite{yue2020interventional} with three rules of \textit{do}-calculus~\cite{pearl1995causal}: Insertion/deletion of observations, Action/observation exchange and Insertion/deletion of actions. For consistency, we describe these three rules as follows~\cite{yue2020interventional}:

Given a causal directed acyclic graph $\mathcal{G}$, denote $X,Y,Z$ and $W$ be arbitrary disjoint sets of nodes. We use $\mathcal{G}_{\overline{X}}$ to denote the manipulated graph where all incoming arrows to node $X$ are deleted. Similarly $\mathcal{G}_{\underline{X}}$ represents the graph where outgoing arrows from node $X$ are deleted. Lower case $x,y,z$ and $w$ denote specific values taken by each set of nodes: $X=x,Y=y,Z=z$ and $W=w$. For any interventional distribution compatible with $\mathcal{G}$, we have the following three rules:

\noindent\textbf{Rule 1} Insertion/deletion of observations:
\begin{equation}
    P(y|do(x),z,w)=P(y|do(x),w), if(Y\Vbar Z|X,W)_{\mathcal{G}_{\overline{X}}}
\end{equation}

\noindent\textbf{Rule 2} Action/observation exchange:
\begin{equation}
\small
    P(y|do(x),do(z),w)=P(y|do(x),z,w), if(Y\Vbar Z|X,W)_{\mathcal{G}_{\overline{X}\underline{Z}}}
\end{equation}

\noindent\textbf{Rule 3} Insertion/deletion of actions:
\begin{equation}
\small
    P(y|do(x),do(z),w)=P(y|do(x),w), if(Y\Vbar Z|X,W)_{\mathcal{G}_{\overline{X}\overline{Z(W)}}},
\end{equation}
where $Z(W)$ is a set of nodes in $Z$ that are not ancestors of any $W$-node in $\mathcal{G}_{\overline{X}}$.

In our causal formulation, the desired interventional distribution $P(Y|do(S=s))$ can be derived by:

\begin{equation}
\begin{aligned}
     P(Y|do(S))&=\sum_{v} P(Y|do(S=s),V=v)P(V=v|do(S=s))\\
     &=\sum_{v} P(Y|do(S=s),V=v)P(V=v)\\
     &=\sum_{v} P(Y|S=s,V=v)P(V=v),
\end{aligned}
\end{equation}
where line 1 follows the law of total probability; line 2 uses Rule 3 given $S\Vbar V$ in $\mathcal{G}_{\overline{X}}$; line 3 uses Rule 2 to change the intervention term to observation as $(Y\Vbar S|V)$ in $\mathcal{G}_{\underline{X}}$.

\ifCLASSOPTIONcaptionsoff
  \newpage
\fi

{\small
\bibliographystyle{IEEEtran}
\bibliography{egbib}
}

\begin{IEEEbiography}[{\includegraphics[width=1in,height=1.25in,clip,keepaspectratio]{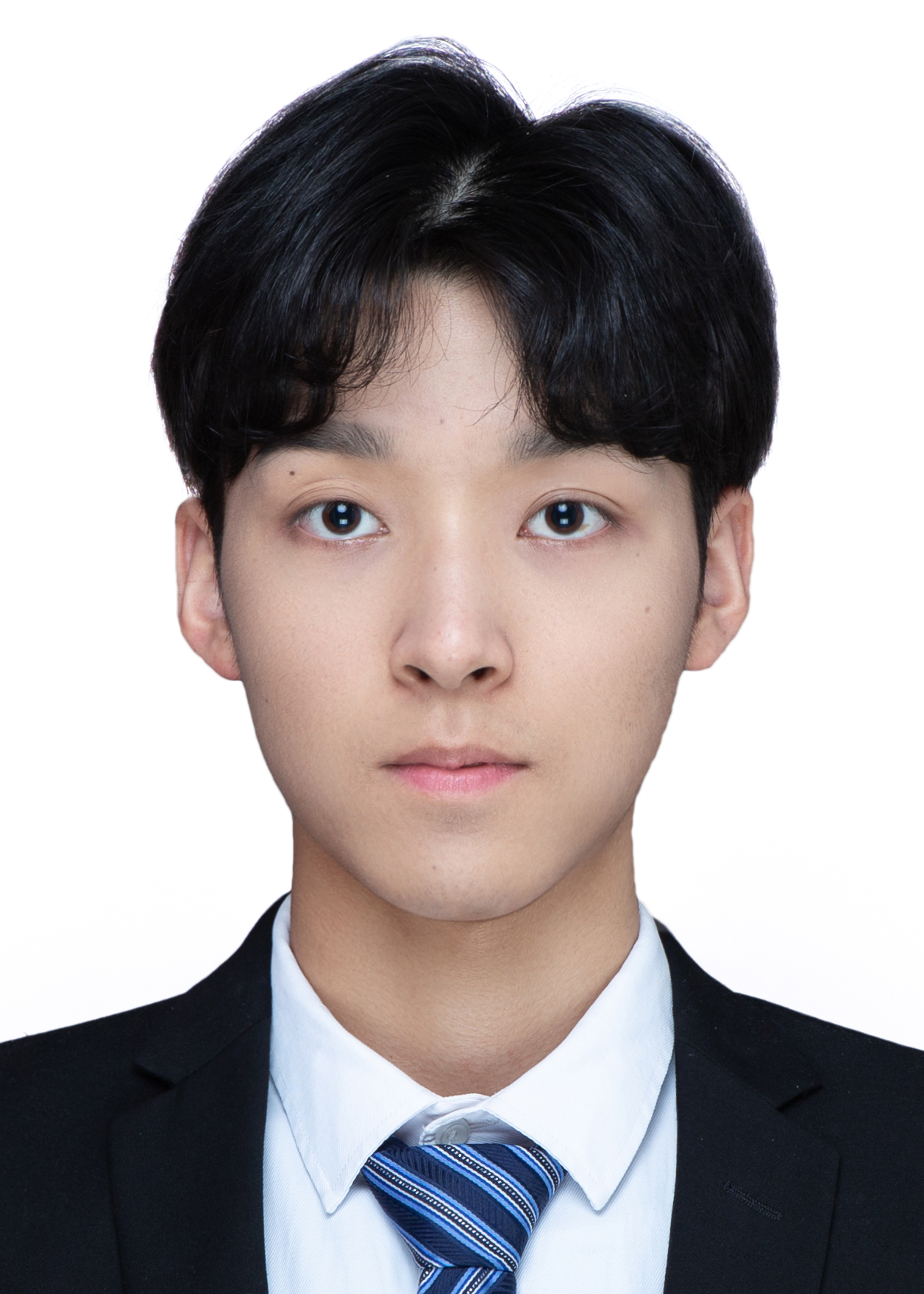}}]{Yi-Fan Zhang} received the B.S. degree from south china university of technology, in 2021. He is currently pursuing the Ph.D. degree with the National Laboratory of Pattern Recognition (NLPR), Institute of Automation, Chinese Academy of Sciences (CASIA), Beijing, China. Her research interests include machine learning, and deep learning
\end{IEEEbiography}


\begin{IEEEbiography}[{\includegraphics[width=1in,height=1.25in,clip,keepaspectratio]{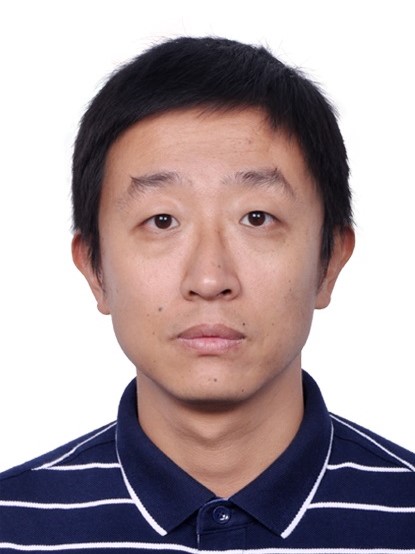}}]{Zhang Zhang}
received the B.S. degree in computer science and technology from Hebei University of Technology, Tianjin, China, in 2002, and the Ph.D. degree in pattern recognition and intelligent systems from the National Laboratory of Pattern Recognition, Institute of Automation, Chinese Academy of Sciences, Beijing, China in 2009. Currently, he is an associate professor at the National Laboratory of Pattern Recognition, Institute of Automation, Chinese Academy of Sciences (CASIA). His research interests include action and activity recognition, human attribute recognition, person re-identification, and large-scale person retrieval. He has published 40s research papers on computer vision and pattern recognition, including some highly ranked journals and conferences, e.g., IEEE TPAMI, IEEE TIP, CVPR, and ECCV.
\end{IEEEbiography}

\begin{IEEEbiography}[{\includegraphics[width=1in,height=1.25in,clip,keepaspectratio]{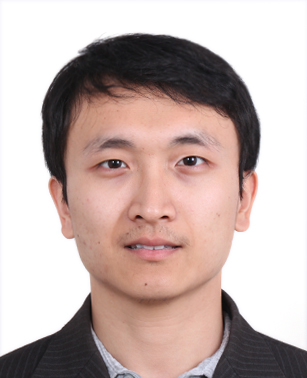}}]{Da Li}
Da Li received the M.Eng. degree in electronics and communication engineering from Suzhou Institute of Nano-Tech and Nano-Bionics, Chinese Academy of Sciences, China, in 2013, and the Ph.D. degree in computer applications technology from the School of Artificial Intelligence, University of Chinese Academy of Sciences, China, in 2020. He is currently a Postdoc with the Center for Research on Intelligent Perception and Computing (CRIPAC), Institute of Automation, Chinese Academy of Sciences. His research interests include big visual data and video surveillance.
\end{IEEEbiography}

\begin{IEEEbiography}[{\includegraphics[width=1in,height=1.25in,clip,keepaspectratio]{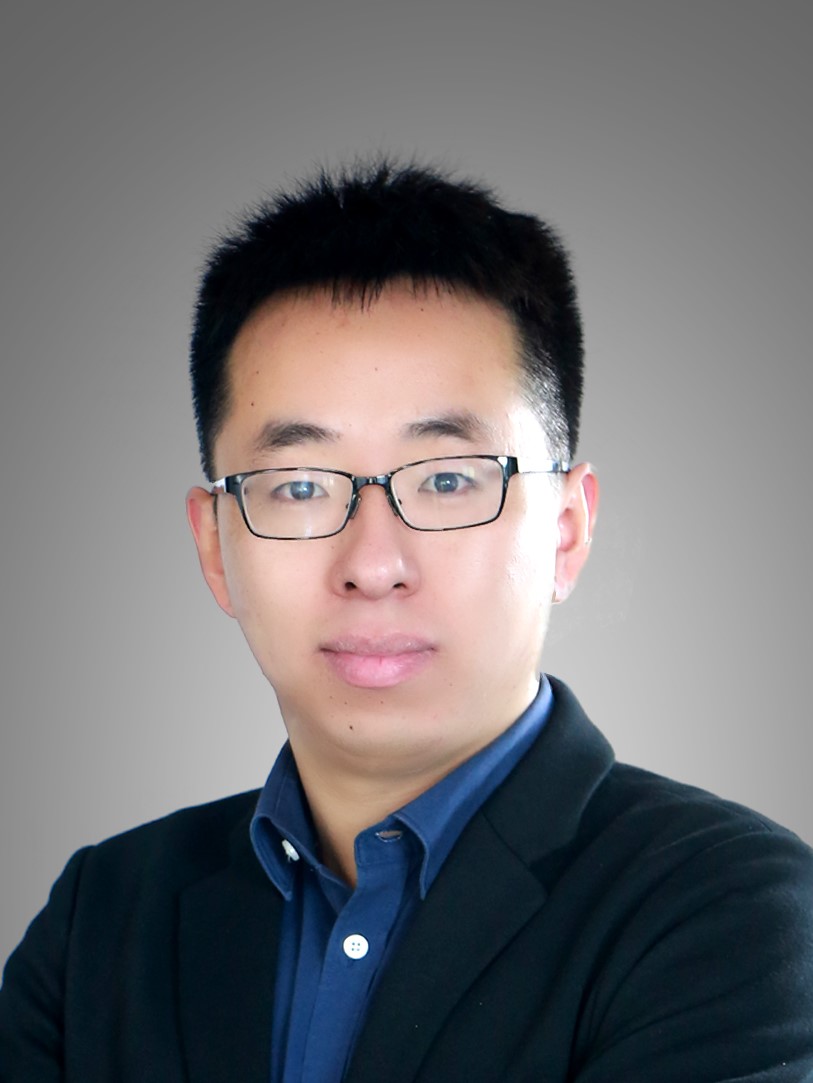}}]{Zhen Jia}
received his B.Eng. degree in electronic engineering from Yunnan University, Kunming, China, in 2013, and the Ph.D. degree in pattern recognition and intelligent systems from the Center for Research on Intelligent Perception and Computing, Institute of Automation, Chinese Academy of Sciences, Beijing, China, in 2020. Currently, he is a postdoctoral researcher at the Center for Research on Intelligent Perception and Computing, Institute of Automation, Chinese Academy of Sciences. His research interests include zero-shot learning, few-shot learning, and generative models.
\end{IEEEbiography}

\begin{IEEEbiography}[{\includegraphics[width=1in,height=1.25in,clip,keepaspectratio]{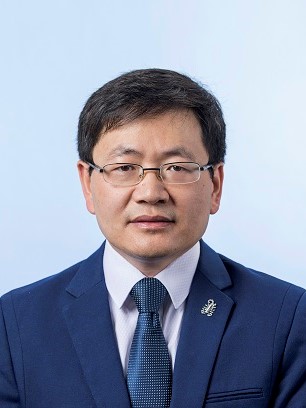}}]{Liang  Wang}
(Fellow, IEEE) received the B.Eng. and M.Eng. degrees from Anhui University in 1997 and 2000, respectively, and the Ph.D. degree from the Institute of Automation, Chinese Academy of Sciences (CASIA) in 2004. From 2004 to 2010, he was a Research Assistant with Imperial College London, U.K., and University, Australia, a Research Fellow with the University of Melbourne, Australia, and a Lecturer with the University of Bath, U.K., respectively. He is currently a Full Professor of the Hundred Talents Program with the National Laboratory of Pattern Recognition, CASIA. His major research interests include machine learning, pattern recognition, and computer vision. He is an IAPR Fellow
\end{IEEEbiography}

\begin{IEEEbiography}[{\includegraphics[width=1in,height=1.25in,clip,keepaspectratio]{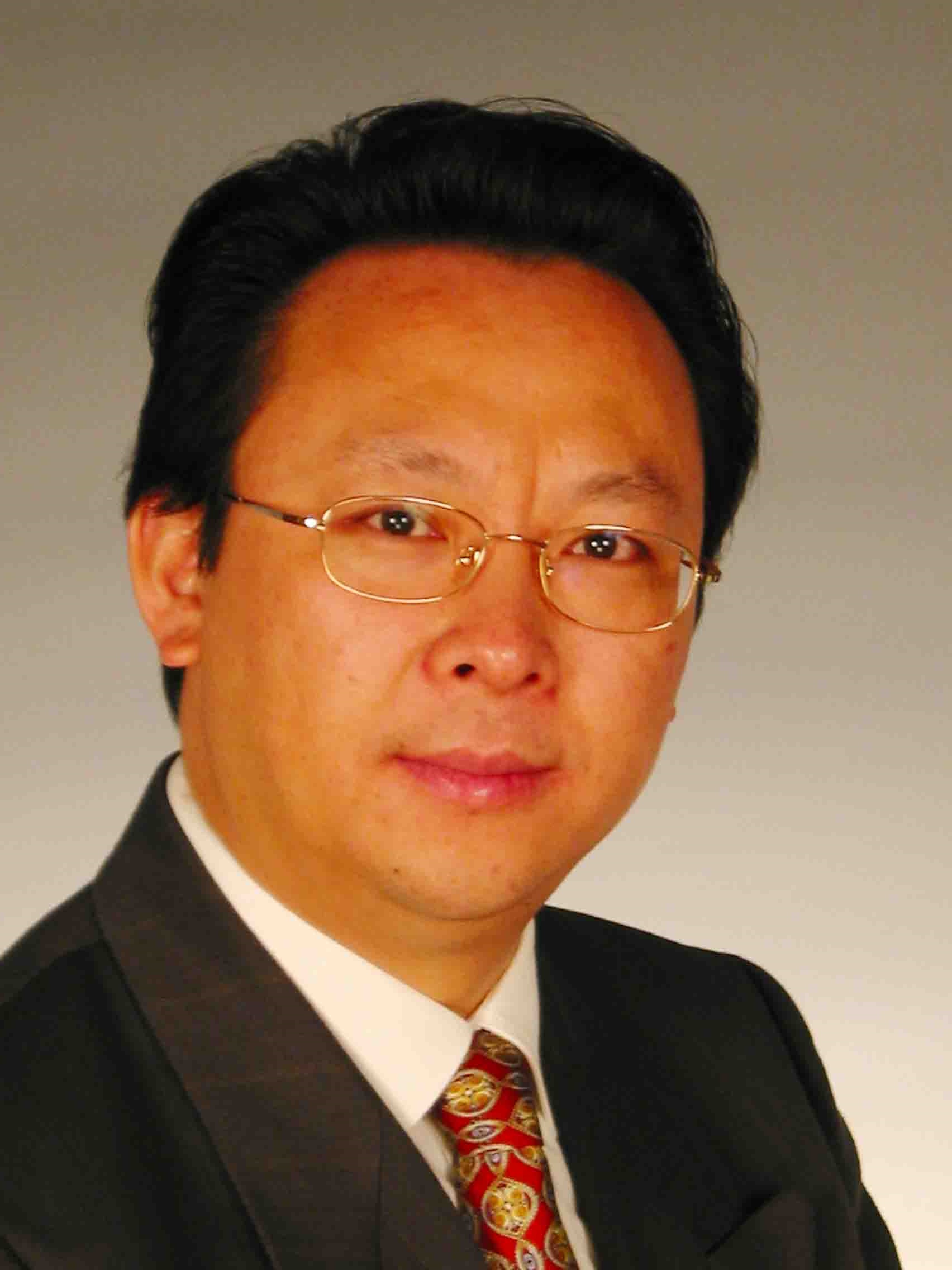}}]{Tieniu Tan}
(Fellow, IEEE) received the B.Sc. degree in electronic engineering from Xi’an Jiaotong University, China, in 1984, and the M.Sc. and Ph.D.degrees in electronic engineering from Imperial College London, U.K., in 1986 and 1989, respectively. He is currently a Professor with Center for Research on Intelligent Perception and Computing, NLPR, CASIA, China. He has published more than450 research papers in refereed international journals and conferences in the areas of image processing,computer vision, and pattern recognition, and has authored or edited 11 books. His research interests include biometrics, image and video understanding, information hiding, and information forensics. He is a Fellow of CAS, TWAS, BAS, IAPR, and the U.K. Royal Academy of Engineering, and the Past President of IEEE Biometrics Council
\end{IEEEbiography}

\end{document}